\documentclass[journal]{IEEEtran}
\UseRawInputEncoding
\usepackage{lineno,hyperref}
\usepackage{graphicx}
\usepackage{epstopdf}
\usepackage{mathrsfs}
\usepackage{amsfonts}
\usepackage{bm}
\usepackage{multirow}
\usepackage{threeparttable}
\usepackage[subfigure]{tocloft}
\usepackage{subfigure}
\ifCLASSINFOpdf
\else
\fi
\begin{document}

\title{Robust and Precise Facial Landmark Detection by Self-Calibrated Pose Attention Network}

\author{Jun~Wan, ~\IEEEmembership{Member,~IEEE,} Hui Xi, Jie~Zhou, ~\IEEEmembership{Member,~IEEE,} Zhihui~lai, ~\IEEEmembership{Member,~IEEE,} Witold~Pedrycz, ~\IEEEmembership{Fellow,~IEEE,} Xu~Wang, Hang~Sun
	
	
	\thanks{This work is supported by the National Natural Science Foundation of China (Grant No. 62002233, 62076164, 61802267, 61976145 and 61806127), the Shenzhen Science and Technology Program (Grant No. JCYJ20210324094601005, JCYJ20210324094413037 and JCYJ20190813100801664), the Natural Science Foundation of Guangdong Province (Grant No. 2019A1515111121, 2021A1515011861) and the Natural Science Foundation of HuBei Province (Grant No. 2021CFB004). Corresponding author: Jie Zhou.}
	\thanks{J. Wan and Hui Xi are with the School of Information and Safety Engineering, Zhongnan University of Economics and Law, Wuhan, 430073, China and the College of Computer Science and Software Engineering, Shen zhen University, Shenzhen, 518060, China. (e-mail:junwan2014@whu.edu.cn, qinghuan\_xi@163.com).}
	\thanks{J. Zhou, Z. Lai and X. Wang are with the Computer Vision Institute, College of Computer Science and Software Engineering, Shenzhen University, Shenzhen 518060, China, and also with the Shenzhen Institute of Artificial Intelligence and Robotics for Society, Shenzhen, China. (e-mail: jie\_jpu@163.com, lai\_zhi\_hui@163.com,  wangxu@szu.edu.cn).}
	\thanks{W. Pedrycz is with the Department of Electrical $\&$ Computer Engineering, University of Alberta, Edmonton, Canada, and the Systems Research Institute, Polish Academy of Sciences, Warsaw, Poland. (e-mail: wpedrycz@ualberta.ca.)}
	\thanks{H. Sun is with College of Computer and Information Technology, China Three Gorges University, Yichang, HuBei, China. (e-mail: sunhang0418@whu.edu.cn.)}%
}

\markboth{Journal of \LaTeX\ Class Files}%
{Shell \MakeLowercase{\textit{et al.}}: Bare Demo of IEEEtran.cls for IEEE Journals}

\maketitle
\begin{abstract}
Current fully-supervised facial landmark detection methods have progressed rapidly and achieved remarkable performance. However, they still suffer when coping with faces under large poses and heavy occlusions for inaccurate facial shape constraints and insufficient labeled training samples. In this paper, we propose a semi-supervised framework, i.e., a Self-Calibrated Pose Attention Network (SCPAN) to achieve more robust and precise facial landmark detection in challenging scenarios. To be specific, a Boundary-Aware Landmark Intensity (BALI) field is proposed to model more effective facial shape constraints by fusing boundary and landmark intensity field information. Moreover, a Self-Calibrated Pose Attention (SCPA) model is designed to provide a self-learned objective function that enforces intermediate supervision without label information by introducing a self-calibrated mechanism and a pose attention mask. We show that by integrating the BALI fields and SCPA model into a novel self-calibrated pose attention network, more facial prior knowledge can be learned and the detection accuracy and robustness of our method for faces with large poses and heavy occlusions have been improved. The experimental results obtained for challenging benchmark datasets demonstrate that our approach outperforms state-of-the-art methods in the literature.

\end{abstract}


\IEEEpeerreviewmaketitle

\section{Introduction}
\IEEEPARstart{F}{acial} landmark detection, also known as face alignment, aims to locate the predefined landmarks (e.g., eye corners, nose tip, mouth corners) of a face, which has attracted much attention from the computer vision community. Precise and robust facial landmark detection lays the foundation for high-quality performance of many computer vision and computer graphics tasks, such as face recognition \cite{ Wang2020ModalRA}, face animation \cite{Wang2020DualLF, Ichim2017PhacePF}, and face reenactment \cite{Ha2020MarioNETteFF, Yao2020MeshGO}. Many face analysis tasks \cite{Xie2019AdaptiveWO, Zhu2020SpatialTemporalKI, Xiong2020ECMLAE} rely on locations of detected facial landmarks, so imprecise landmarks will be propagated to the subsequent tasks and could lead to unsatisfactory face analysis results.
\begin{figure}[!t]
	\begin{center}
		\includegraphics[width=0.8\linewidth]{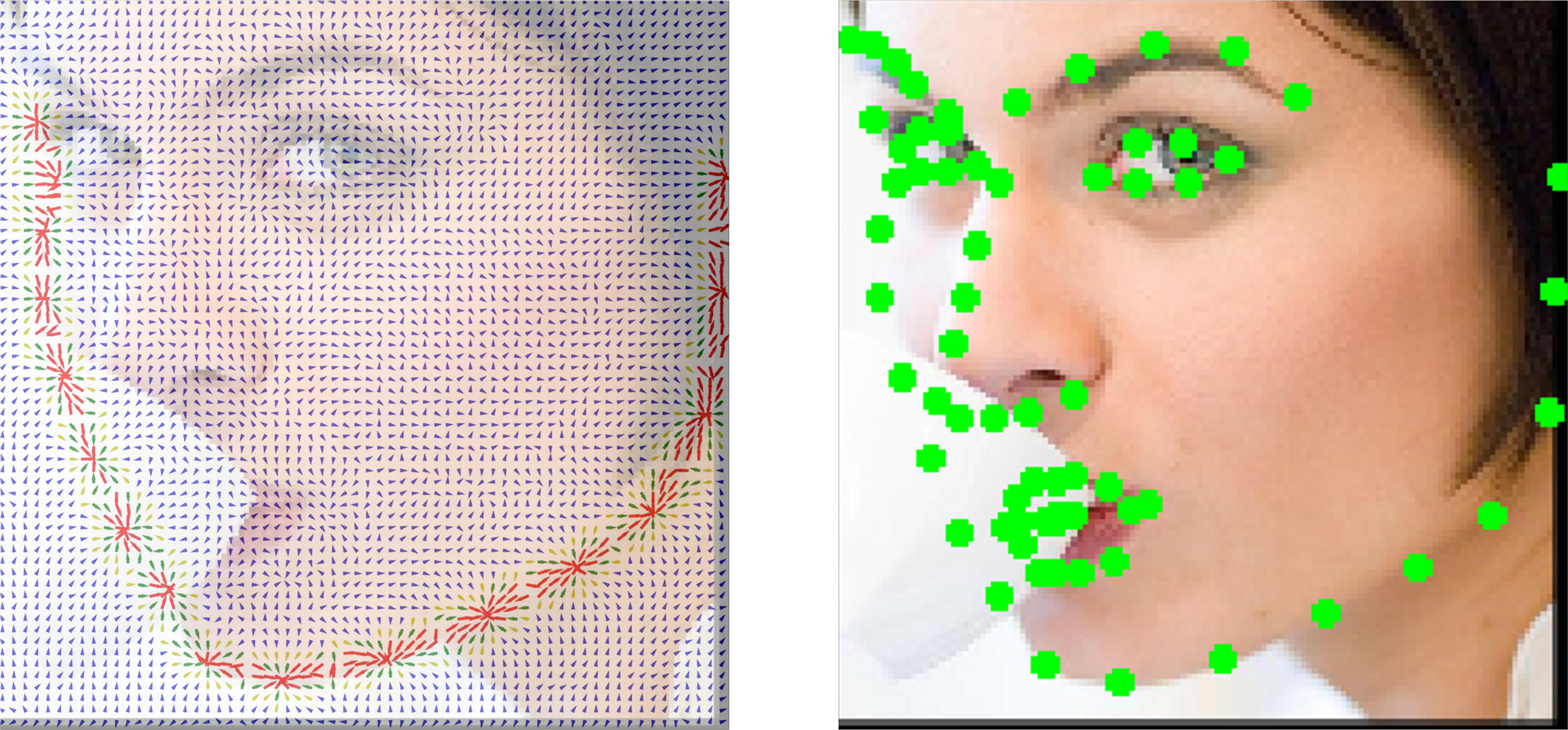}
	\end{center}
	\caption{The proposed Boundary-Aware Landmark Intensity (BALI) fields. By integrating the facial boundary and landmark intensity field information, both boundary and field constraints can be introduced to achieve more robust and precise facial landmark detection.}
	\label{problem}
\end{figure}

Most works \cite{Sukno20153DFL, 2016Automatic, lin2019task, Kumar2020LUVLiFA, Wan2021RobustFA} of facial landmark detection adopt the supervised learning approach and they usually map facial appearances to landmark heatmaps or coordinates, which have achieved great success. However, on the one hand, their performance depends on a large number of training samples with full landmark annotations that are usually tedious and time-consuming to annotate. For example, for 3000 images and 68 landmarks a face, 204000 landmarks need to be annotated. Moreover, the limitations of the human visual system also reduce the precision and consistency of landmarks. On the other hand, most facial landmark detection methods suffer from performance degradation when facing large poses and partial occlusions, as in this case, convolutional neural networks (CNNs) may be misled to learn inaccurate feature representation and shape constraints. Facial boundary heatmaps \cite{Wu2018LookAB, Wan2021RobustFL} and part affinity fields \cite{Cao2017RealtimeM2, Cao2021OpenPoseRM} are proposed to address these problems. However, the constraints of both boundary heatmaps and part affinity fields are very rough, and it is still hard to achieve high-precision landmark detection. Therefore, how to model more effective facial shape constraints for precise and robust landmark detection with unlabeled face images remains a challenging problem. 

\begin{figure*}[!t]
  \centering
  \includegraphics[width=0.98\linewidth]{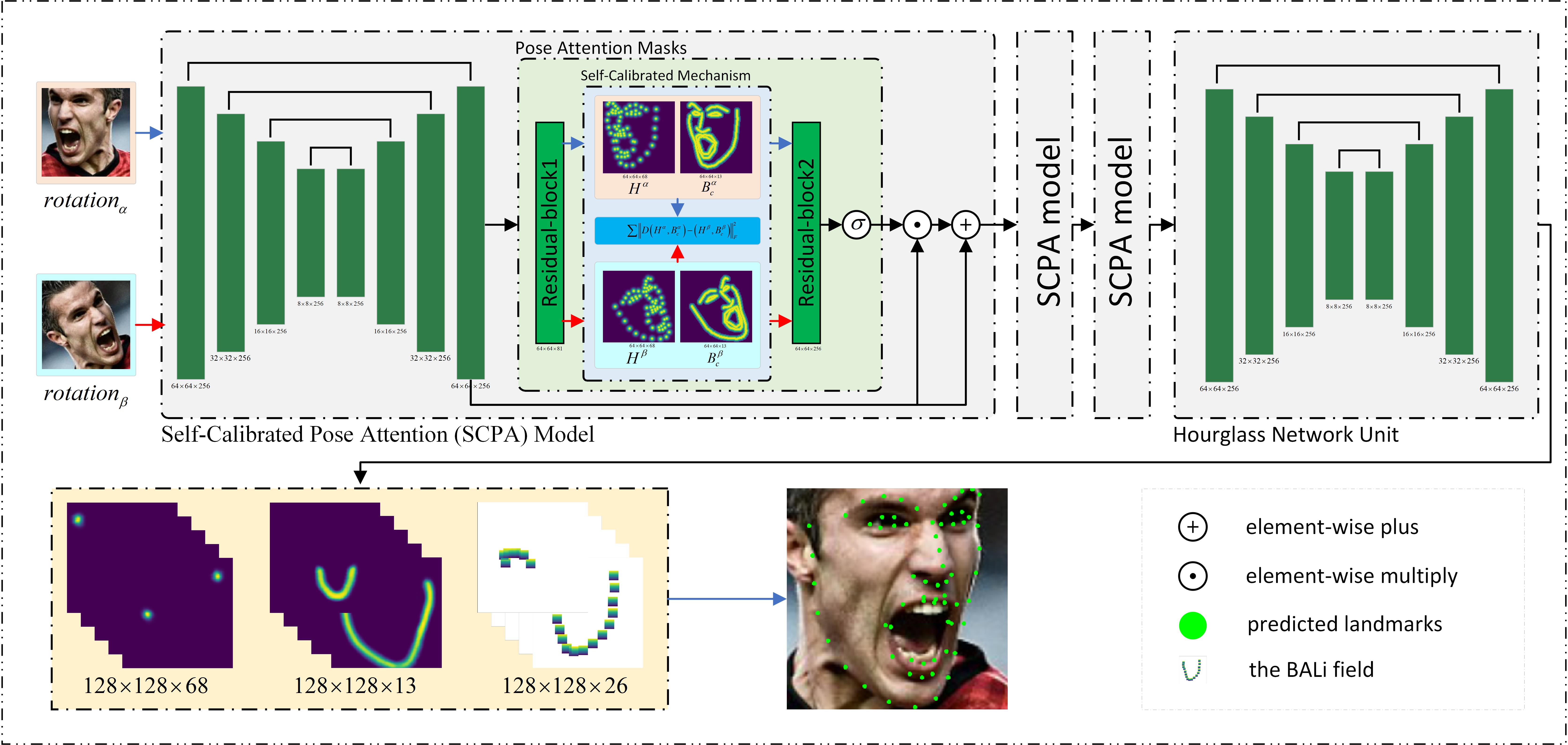}
  \centering
  \caption{The overall architecture of the proposed Self-Calibrated Pose Attention Network (SCPAN). The proposed BALI fields can introduce both boundary and field constraints to the predicted landmarks and the SCPA model is designed to learn more representative and discriminative features by incorporating a self-calibrated mechanism and a pose attention mask. Then, by integrating the BALI fields and SCPA model into a self-calibrated pose attention network via a seamless formulation, more effective landmark heatmaps and BALI fields can be produced, thus achieving robust and precise facial landmark detection.}
  \label{SCPAN}
\end{figure*}

To address the above problems, in this paper, we propose a novel semi-supervised approach, i.e., Self-Calibrated Pose Attention Network (SCPAN) for achieving robust and precise facial landmark detection. The overall architecture of the proposed SCPAN is shown in Fig. \ref{SCPAN}. SCPAN contains two parts: Boundary-Aware Landmark Intensity (BALI) fields and Self-Calibrated Pose Attention (SCPA) model. The proposed BALI fields have a composite structure, which is composed of a scalar component for the confidence of a particular boundary and a vector component that points to the closest landmark in the particular boundary. As shown in Fig. \ref{problem}, the proposed BALI fields can simultaneously add both boundary and field constraints to the predicted landmarks, thus helping improve the detection accuracy. Moreover, an SCPA model is designed to learn more representative and discriminative features by introducing the self-calibrated mechanism and pose attention mask. The self-calibrated mechanism can provide a natural learning objective function that enforces intermediate supervision without label information, which effectively reduces the dependence of the detection accuracy on labeled facial images. The pose attention mask can selectively emphasize important features and suppress less useful ones for producing more effective landmark heatmaps and BALI fields. Finally, by integrating the proposed BALI fields and SCPA model into a novel SCPAN framework with seamless formulations, more facial prior knowledge can be learned for achieving robust and precise facial landmark detection. The main contributions of this work are summarized as follows:

1) By incorporating boundary heatmaps and landmark intensity fields, we propose a Boundary-Aware Landmark Intensity (BALI) field, in which both boundary and field information can be better used to model the facial shape constraints for detecting more accurate landmarks.

2) A Self-Calibrated Pose Attention (SCPA) model is proposed to learn more representative and discriminative features by introducing the self-calibrated mechanism and pose attention mask. It can help generate more effective landmark heatmaps and BALI fields while dealing with complicated cases, especially for faces with large poses and heavy occlusions. 

3) To the best of our knowledge, this is the first study to explore how to incorporate landmark heatmaps, boundary heatmaps and landmark intensity fields for handling facial landmark detection under challenging scenarios in a semi-supervised way. By seamlessly integrating BALI fields and SCPA model, the proposed SCPAN outperforms state-of-the-art methods on challenging benchmark datasets such as 300W \cite{Sagonas2016300FI}, Menpo 2D \cite{Deng2018TheMB}, COFW \cite{Burgosartizzu2013Robust}, AFLW \cite{Zhu2016UnconstrainedFA}, WFLW \cite{Wu2018LookAB} and 300VW \cite{Shen2015TheFF}.

The rest of the paper is organized as follows. Section \textbf{II} gives an overview of the related work. Section \textbf{III} shows the proposed method, including the BALI fields and the SCPA model. A series of experiments are conducted to evaluate the performance of the proposed method in Section \textbf{IV}. Finally, Section \textbf{V} concludes the paper.
\section{Related Work}
This section reviews related work completed for fully-supervised facial landmark detection and semi-supervised facial landmark detection methods.

\textbf{Fully-Supervised Facial Landmark Detection. } The mainstream of fully-supervised facial landmark detection usually maps facial appearance features to landmark coordinates or heatmaps and has achieved great success. Early methods such as Active Shape Model (ASM) \cite{cootes1995active}, Active Appearance Model(AAM) \cite{cootes2001active} and Constrained Local Model (CLM) \cite{Cristinacce2006FeatureDA}, use the parametric models to enhance the shape variation. They are sensitive to variations on facial poses and occlusions. Recent approaches can be divided into two groups: coordinate regression-based and heatmap regression-based methods. The coordinate regression-based methods\cite{cao2014face, ren2014face, Trigeorgis2016MnemonicDM, Wu2018LookAB} directly learn the mapping from facial appearance feature to the landmark coordinate vectors by using different models. In Mnemonic Descent Method (MDM) \cite{Trigeorgis2016MnemonicDM}, a recurrent neural network is used to extract the task-based features and model dependencies between cascade iterations for detecting more accurate landmarks. In look-at-boundary (LAB) \cite{Wu2018LookAB}, the stacked hourglass network is used to generate more effective facial boundary heatmaps by introducing the adversarial concept and message passing layers, which helps enhance the shape constraints and improve alignment accuracy. In the occlusion-adaptive deep network (ODN) \cite{Zhu2019RobustFL}, the Resnet is used to address the occlusion problem for facial landmark detection by recovering more discriminative representations with the learned geometric information. With their favorable regression abilities, these algorithms all achieve good results in restricted conditions. However, they usually regress landmark coordinates by using full connection operations, which can not fully utilize the spatial relationships between pixels and limits their accuracy and robustness against faces in the wild. Since the heatmap regression-based face alignment methods \cite{Dong2018StyleAN, Liu2019SemanticAF, Wan2021RobustFA, Wan2021RobustFL, Kumar2020LUVLiFA} predict landmarks by regressing landmark heatmaps, which makes them better encode the part constraints and context information, thus achieving state-of-the-art performance. Dong et al. propose a style aggregated network (SAN) \cite{Dong2018StyleAN} to address face alignment problems under image styles variations. Liu et al. \cite{Liu2019SemanticAF} propose a novel latent variable optimization strategy to find the semantically consistent annotations and alleviate the limitations of human annotations. In Multi-Order High-Precision Hourglass Network (MMHN) \cite{Wan2021RobustFA} and Multi-order Multi-constraint Deep Networks (MMDN) \cite{Wan2021RobustFL}, high-order information are utilized to explore more discriminative representations for robust face alignment. In LUVLi \cite{Kumar2020LUVLiFA}, a novel end-to-end framework is proposed to achieve state-of-the-art alignment accuracy by jointly estimating facial landmark locations, uncertainty, and visibility. However, their performance still depends on a large scale of annotated training samples and also suffers from faces with large poses and heavy occlusions. While our proposed SCPAN can model more effective facial shape constraints with unlabeled face images, thus reducing the dependency on landmark annotations.

\textbf{Semi-Supervised Facial Landmark Detection. } As the fully-supervised facial landmark detection methods \cite{Liu2019SemanticAF, Wan2021RobustFA, Wan2021RobustFL, Kumar2020LUVLiFA} depend highly on the scale of annotated face images, several semi-supervised methods \cite{Tang2018FacialLD, Honari2018ImprovingLL, Dong2018SupervisionbyRegistrationAU, Zhu2020SpatialTemporalKI} are proposed to improve face alignment by using unlabeled face images. Tang et al. \cite{Tang2018FacialLD} use an iterative coarse-to-fine patch-based scheme and a greedy patch selection strategy to address face alignment by optimizing an objective function defined on both annotated and unannotated images. Honari et al. \cite{Honari2018ImprovingLL} aim to improve face alignment by proposing an unsupervised technique that leverages equivariant landmark transformation and auxiliary attributes, thereby reducing the need of labeled face images. Dong et al. \cite{Dong2018SupervisionbyRegistrationAU} use the coherency of optical flow as the source of supervision, which can help achieve more precise facial landmark detection. Dong et al. \cite{Dong2019TeacherSS} propose an interaction mechanism between a teacher and two students to generate more reliable pseudo labels for addressing partially labeled facial landmark detection problems. Zhu et al. \cite{Zhu2020SpatialTemporalKI} and Yin et al. \cite{Yin2020ExploitingSA} both use the consistency constraints on the facial sequence to address semi-supervised facial landmark tracking tasks. However, utilizing optical flow or the temporal relations can only address face alignment problems under small facial poses and expressions variations, as an ``in-the-wild" facial video usually deforms or zooms gradually and smoothly without sharp changes, thus resulting in a decrease in detection accuracy when facing large variations on facial poses and expressions. By contrast, our SCPAN is able to use unlabeled face images with large poses and heavy occlusions as supervision signals, therefore its detection robustness and accuracy have been enhanced.
\section{Robust and Precise Facial Landmark Detection by Self-Calibrated Pose Attention Network}
In this section, we first elaborate on the Boundary-Aware Landmark Intensity (BALI) fields, and then describe the Self-Calibrated Pose Attention (SCPA) model. Finally, we show the proposed Self-Calibrated Pose Attention Network (SCPAN) and its objective function.
\begin{figure}[t]
	\begin{center}
		\includegraphics[width=0.96\linewidth]{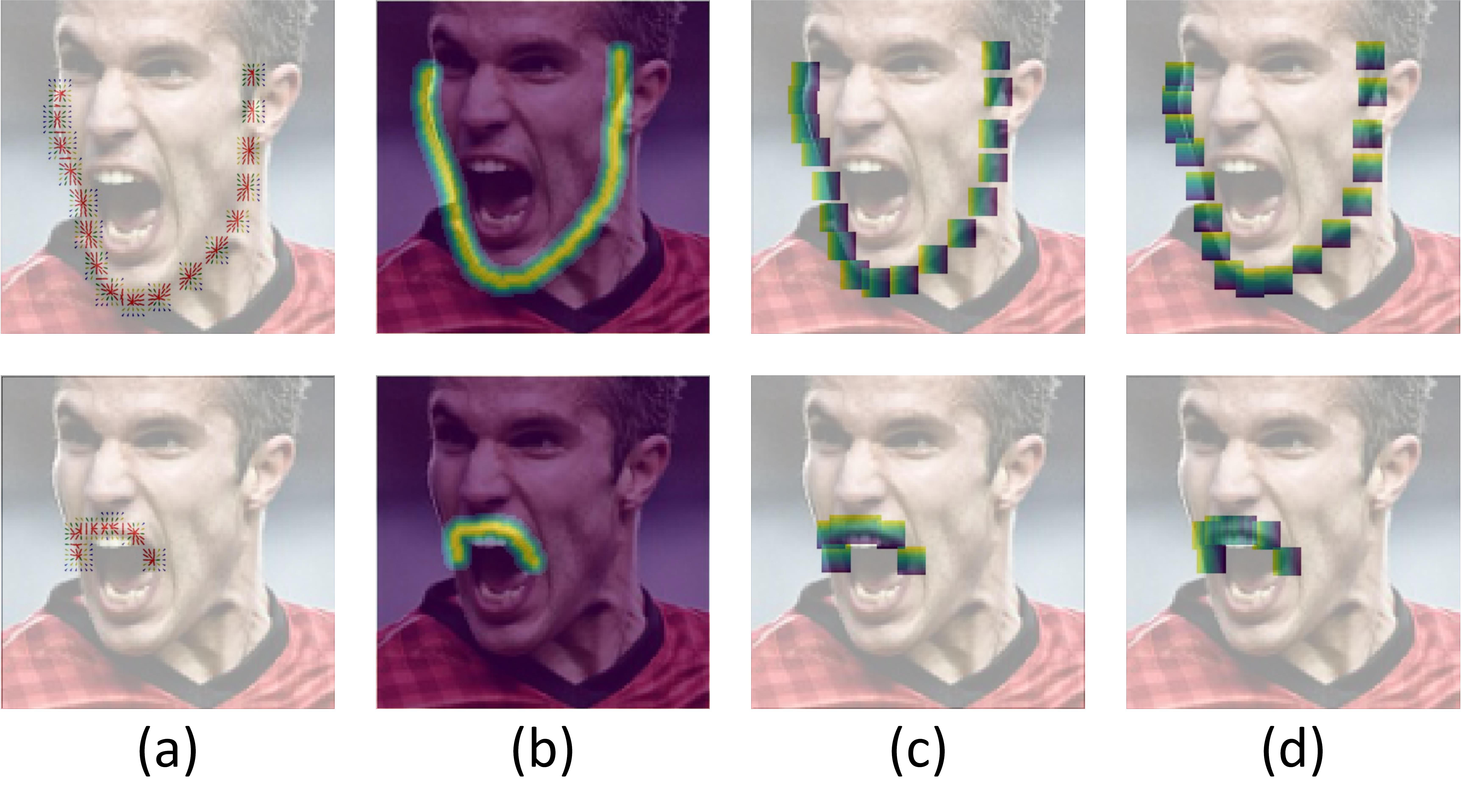}
	\end{center}
	\caption{(a) The proposed Boundary-Aware Landmark Intensity (BALI) fields. (b) boundary heatmaps. (c) the x-offset of BALI field. (d) the y-offset of BALI field. The proposed BALI fields can introduce both boundary and field information, which helps better model facial shape constraints for faces with large poses and heavy occlusions. }
	\label{fig3}
\end{figure}
\subsection{Boundary-Aware Landmark Intensity (BALI) fields}
The proposed BALI fields have a composite structure. They consist of a scalar component for confidence (boundary heatmaps as shown in Fig. \ref{fig3} (b)) and a vector component that points to the closest landmark in this boundary (as shown in Fig. \ref{fig3} (c) and (d)). The proposed BALI fields need to estimate the confidence $B^{ij}_c$ and a vector $(B_u^{ij}, B_v^{ij})$ at every output location $(i, j)$, which can be expressed as follows:

\begin{equation}
B^{ij}=\left\{{B^{ij}_c, B^{ij}_{ u}, B^{ij}_{v}} \right\}
\end{equation}
\begin{equation}
(B_u^{ij}, B_v^{ij})=(\hat u, \hat v)-(i, j)
\end{equation}where $(B_u^{ij}, B_v^{ij})$ is calculated within a square region with edge length $2R+1$ and centered at the ground-truth landmark location $(\hat u, \hat v)$. $B_c^{ij}$ denotes the boundary heatmap which is constructed according to MMDN \cite{Wan2021RobustFL}. To be specific, for each boundary, landmarks on this boundary are firstly interpolated to get a dense boundary line. Then, a Gaussian distribution is used to construct a ground-truth boundary heatmap by transforming distance map $Dist$ that is obtained with a binary map and distance transform function. The boundary heatmap is constructed as follows:

\begin{small}
\begin{equation}
		\hat{B}_c^{ij} = \left\{ {\begin{array}{*{20}{c}}
				{\exp \left( { - \frac{{Dist\left( {i, j} \right)}}{{2\sigma ^2}}} \right){\kern 1pt} {\kern 1pt} {\kern 1pt} {\kern 1pt} {\kern 1pt} {\kern 1pt} ,if{\kern 1pt} Dist\left( {i,j} \right) < 2{\sigma}{\kern 1pt} }\\
				{\xi ,{\kern 1pt} {\kern 1pt} {\kern 1pt} {\kern 1pt} {\kern 1pt} {\kern 1pt} {\kern 1pt} {\kern 1pt} {\kern 1pt} {\kern 1pt} {\kern 1pt} {\kern 1pt} {\kern 1pt} {\kern 1pt} {\kern 1pt} {\kern 1pt} {\kern 1pt} {\kern 1pt} {\kern 1pt} {\kern 1pt} {\kern 1pt} {\kern 1pt} {\kern 1pt} {\kern 1pt} {\kern 1pt} {\kern 1pt} {\kern 1pt} {\kern 1pt} {\kern 1pt} {\kern 1pt} {\kern 1pt} {\kern 1pt} {\kern 1pt} {\kern 1pt} {\kern 1pt} {\kern 1pt} {\kern 1pt} {\kern 1pt} {\kern 1pt} {\kern 1pt} {\kern 1pt} {\kern 1pt} {\kern 1pt} {\kern 1pt} {\kern 1pt} {\kern 1pt} {\kern 1pt} {\kern 1pt} {\kern 1pt} {\kern 1pt} {\kern 1pt} {\kern 1pt} {\kern 1pt} {\kern 1pt} {\kern 1pt} {\kern 1pt} {\kern 1pt} {\kern 1pt} {\kern 1pt} {\kern 1pt} {\kern 1pt} {\kern 1pt} {\kern 1pt} {\kern 1pt} {\kern 1pt} {\kern 1pt} {\kern 1pt} {\kern 1pt} {\kern 1pt} {\kern 1pt} {\kern 1pt} {\kern 1pt} {\kern 1pt} {\kern 1pt} {\kern 1pt} {\kern 1pt} {\kern 1pt} {\kern 1pt} {\kern 1pt} {\kern 1pt} {\kern 1pt} {\kern 1pt} {\kern 1pt} {\kern 1pt} {\kern 1pt} {\kern 1pt} {\kern 1pt} {\kern 1pt} {\kern 1pt} {\kern 1pt} {\kern 1pt} {\kern 1pt} {\kern 1pt} {\kern 1pt} {\kern 1pt} {\kern 1pt} {\kern 1pt} {\kern 1pt} {\kern 1pt} {\kern 1pt} {\kern 1pt} otherwise}
		\end{array}} \right.
\end{equation}
\end{small}where $\xi $ is a small constant, $\sigma$ denotes the standard deviation of the corresponding Gaussian distribution and $\hat{B}$ denotes the ground-truth boundary heatmap. So far, the BALI fields have been constructed, however, how to utilize the BALI fields to detect more accurate landmarks is still an unsolved problem. In this paper, we address the above problem with the help of landmark heatmaps, i.e., we generate the landmark heatmap and the BALI fields at the same time. Therefore, by optimizing landmark heatmaps and boundary heatmaps (contained in the proposed BALI fields) in a multi-task way, the facial boundary constraints can be introduced to generate more effective landmark heatmaps. Then, we further use the field constraints to obtain more precise landmark coordinates. To be specific, we firstly compute the coarse landmark coordinates by using $\arg \mathop {\max }\limits_{i,j} \left( H \right) = \left( {i',j'} \right)$, where $H$ denotes the predicted landmark heatmap. Then, we crop a small square region from the landmark heatmap with edge length $2r+1$ center at $(i', j')$, which is denoted by $H'$. Finally, the soft-argmax operation is utilized on $H'$ to calculate the final landmark coordinates. The whole process can be formulated as follows:

\begin{equation}
H' = \left\{ {\left( {i,j} \right)\left| {i \in \left[ {i' - r,i' + r} \right],} \right.j \in \left[ {j' - r,j' + r} \right]} \right\}
\end{equation}
\begin{equation}
\left( {\bar u, \bar v} \right) = \frac{{\sum\nolimits_{i,j \in H'} {{H^{ij}} \times \left( {\left( {B_u^{ij},B_v^{ij}} \right) + \left( {i,j} \right)} \right)} }}{{\sum\nolimits_{i,j \in H'} {{H^{ij}}} }}
\end{equation}where $(\bar u, \bar v)$ denotes the final predicted landmark coordinates. Therefore, by designing Eq. (5), both boundary and field constraints can be introduced to detect more accurate landmarks.
\subsection{Self-Calibrated Pose Attention model}
The proposed Boundary-Aware Landmark Intensity (BALI) fields can introduce both boundary and field constraints that would be helpful for detecting more precise landmarks, however, how to generate more accurate and effective BALI fields and landmark heatmaps are still open questions. The heatmap regression-based facial landmark detection methods \cite{Dong2018StyleAN, Liu2019SemanticAF, Kumar2020LUVLiFA, Wan2021RobustFA, Wan2021RobustFL} can generate effective landmark heatmaps and have achieved state-of-the-art performance as they can effectively encode the part constraints and context information. However, these methods suffer from performance degradation when facing large poses and heavy occlusions, because 1) the large pose or occlusions will mislead models' learning of robust features and accurate facial shape constraints. 2) backpropagated gradients diminish in strength as they are propagated through a very deep network. Therefore, in this paper, a Self-Calibrated Pose Attention (SCPA) model is proposed to address the above problem by introducing a self-calibrated mechanism and a pose attention mask. The self-calibrated mechanism is able to provide intermediate supervision and address the gradient vanishing problem by optimizing the $L_2$-loss of the generated heatmaps between paired images, and the pose attention mask can drive the network to focus on part of interest by using the newly generated heatmaps as attention for learning more representative and discriminative features. Therefore, the SCPA model is able to produce more accurate BALI fields and landmark heatmaps and help achieve more robust and precise facial landmark detection. The network structure of the proposed SCPA model is shown in Fig. \ref{SCPAN}.
\subsubsection{Self-Calibrated Mechanism}
In CPM \cite{Wei2016ConvolutionalPM} and Openpose \cite{Cao2017RealtimeM2}, the sequential architecture is utilized to learn more effective human pose estimation models by communicating increasingly refined heatmaps between stages and providing a natural learning objective function for enforcing intermediate supervision. Therefore, the problem of vanishing gradients during training can be well addressed. Inspired by this observation, we generate and optimize landmark and boundary heatmaps in each SCPA model, and the loss of the generated landmark and boundary heatmaps is used to train the whole network. 
\begin{figure}[t]
	\begin{center}
		\includegraphics[width=0.88\linewidth]{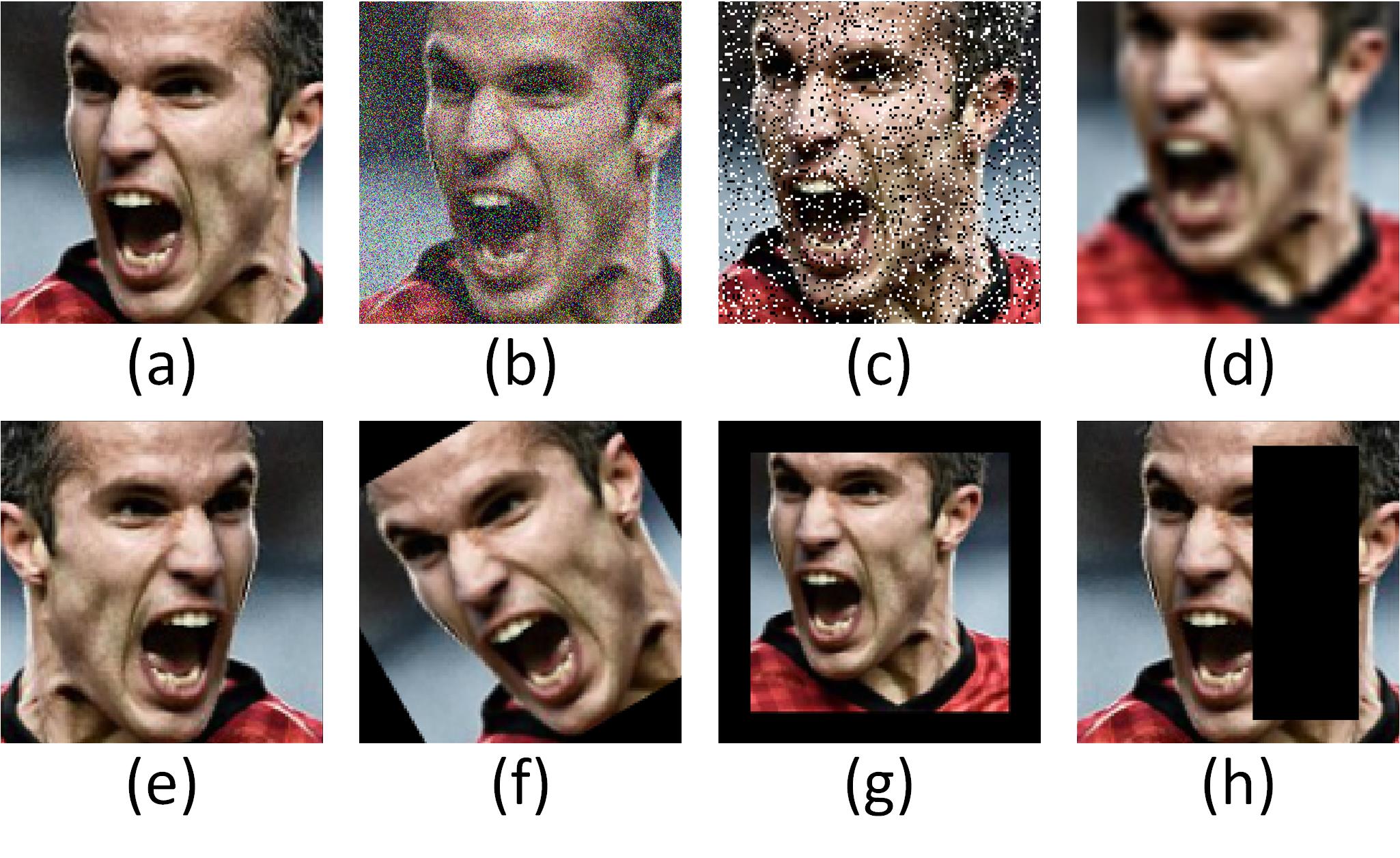}
	\end{center}
	\caption{Visualization of paired images. (a) The original face image. (b) Gaussian noise. (c) Salt noise. (d) Blur. (e) Mirror flip. (f) Rotation. (g) Shape scaling. (h) Occluded by black. With these disturbance operations, the original label information can be fully utilized to achieve more robust and precise landmark detection for faces with larges poses and heavy occlusions.}
	\label{fig5}
\end{figure}
As shown in Fig. \ref{SCPAN}, the SCPA model is actually a modified Hourglass Network unit \cite{Yang2017StackedHN}. The input of the Hourglass Network Unit is denoted as $P$, and the output is denoted as $Q$. To produce landmark and boundary heatmaps, $Q$ will go through a residual-block, and the loss between the generated landmark and boundary heatmaps and the ground-truths is usually used to optimize the SCPA model. However, that loss is calculated based on the training samples' labels, which are usually tedious and time-consuming to obtain. Hence, we further propose a new self-calibrated mechanism, in which the loss of the generated landmark and boundary heatmaps between paired images (called \textbf{self-calibrated loss}) is introduced as part of loss for supervision. On the one hand, the proposed self-calibrated mechanism can boost supervision by introducing the self-calibrated loss. On the other hand, the calculation of self-calibrated loss may not need the label information that can effectively reduce the dependence of labeled facial images. Paired images mean the original image and its disturbed version. As shown in Fig. \ref{fig5}, to build a disturbed image, a combination of the texture disturbance operations and spatial transformation operations are applied to the original image. The texture disturbance can be achieved by the operations such as occlusion, blurring and noises, while the spatial transformation operations can be implemented by the translation, rotation and scaling operation. To be specific, by inputting a pair of face images $(I^{\alpha}, I^{\beta})$ (the original face $I^{\alpha}$ and its disturbance $I^{\beta}$), the SCPA model produces the corresponding landmark heatmaps $(H^{\alpha}, H^{\beta})$ and boundary heatmaps $(B_c^{\alpha}, B_c^{\beta}) $. Hence, the original loss function can be formulated as follows:

\begin{small}
\begin{equation}
{\mathbb{L}_{org}} = \sum\limits_{n = 1}^N {\sum\limits_{k \in \{ \alpha ,\beta \} } {\left\| {\left( {{H^{n,k}},B_c^{n,k}} \right) - \left( {{{\hat H}^{n,k}},\hat B_c^{n,k}} \right)} \right\|_F^2} }
\end{equation}
\end{small}where $n$ denotes the image index and ${\left\| {{\kern 1pt} {\kern 1pt} {\kern 1pt} {\kern 1pt}  \cdot {\kern 1pt} {\kern 1pt} {\kern 1pt} } \right\|_{\rm{F}}}$ is the Frobenius norm. $\alpha$ and $\beta$ corresponds to the original face and its disturbance, respectively. $(\hat H, \hat B_c)$ represents the ground-truth landmark and boundary heatmaps. Then, the self-calibrated loss can be formulated as follows:

\begin{small}
	\begin{equation}
		{\mathbb{L}_{scl}} = \sum\limits_{n = 1}^N {\left\| {D\left( {{H^{n,\alpha }},B_c^{n,\alpha }} \right) - \left( {{H^{n,\beta }},B_c^{n,\beta }} \right)} \right\|_F^2}
	\end{equation}
\end{small}where $D$ corresponds to the disturbance operations. For the texture disturbance, the heatmaps of paired images are the same, $D$ is equal to 1. For the spatial transformation operations, $D$ is the corresponding transformation parameter. $L_{scl}$ represents the self-calibrated loss. The final loss function corresponding to the self-calibrated mechanism can be expressed as follows:

\begin{small}
	\begin{equation}
	{\mathbb{L}_{scm}} = \eta {\mathbb{L}_{scl}} + \lambda {\mathbb{L}_{org}}
	\end{equation}
\end{small}where $\lambda$ and $\eta$ correspond to the weights of $\mathbb{L}_{org}$ and ${\mathbb{L}_{scl}}$, respectively. More importantly, when $\lambda$ is set to 0, the proposed self-calibrated mechanism is able to use paired unlabeled images as supervision signals and reduce the dependence of label information. This means the detection accuracy of the proposed method is able to be further boosted by using unlabeled face data.
\subsubsection{Pose Attention Mask}
Since the proposed self-calibrated mechanism can help produce more accurate and effective landmark and boundary heatmaps, which contain rich facial pose information that would be helpful for learning more discriminative representations. Therefore, the learned facial pose information can be leveraged to guide the model by hinting where needs to be noticed and where can be ignored. In the SCPA model, such hints are realized by the pose attention mask denoted as $M_t$, which are computed from the pose information that incorporating both landmark and boundary heatmaps. To be specific, the pose information goes through a residual block and an element-wise sigmoid function, and its values are between 0 and 1 indicating the importance of each element in the pose attention mask. The whole process can be formulated as follows:

\begin{small}
	\begin{equation}
	{M_t} = sig \left( {R{B_2}\left( {R{B_1}\left( {{Q_{t}}} \right)} \right)} \right)
	\end{equation}
\end{small}where $M_t$ denotes the pose attention mask in stage $t$ and $t = 1 \cdots 3$. $RB$ represents a residual block and $sig$ denotes the sigmoid function. Having computed $M_t$, the intput of next SPCA model is updated by:

\begin{small}
	\begin{equation}
	{P_{t+1}} = sig \left( {R{B_2}\left( {R{B_1}\left( {{Q_{t}}} \right)} \right)} \right)\odot {Q_{t}} + {Q_{t}}
	\end{equation}
\end{small}where $\odot$ denotes element-wise product. By multiplying $Q_{t}$ with the attention masks $M_t$, the input of the next SPCA model (i.e., $P_{t+1}$) at certain locations are either preserved or suppressed. Since the self-calibrated mechanism can boost supervision of the network and the pose attention mask can selectively emphasize important features and suppress less useful ones, the proposed SPCA model is able to learn more representative and discriminative features for detecting more accurate landmarks.
\subsection{Self-Calibrated Pose Attention Network}
The proposed BALI field is able to achieve highly precise landmark detection by modeling both boundary and field constraints. Then, the SCPA model can learn more representative and discriminative features by introducing the self-calibrated mechanism and the pose attention mask. Finally, by integrating the SCPA model and BALI fields into a Self-Calibrated Pose Attention Network (SCPAN), we can generate more accurate and effective landmark heatmaps and boundary-aware landmark intensity fields for achieving more robust and precise facial landmark detection. The overall network structure of the proposed SCPAN is shown in Fig. \ref{SCPAN}. 
\subsection{Objective Function}
The proposed SCPAN outputs landmark heatmaps and BALI fields, therefore, the loss between the generated landmark heatmaps and BALI fields and the ground-truths should be used as part of the objective function. Moreover, the loss between the paired images and the loss between the predicted landmark coordinates and the ground-truth coordinates should be a part of the final objective function. Therefore, the objective function can be formulated as follows:

\begin{small}
	\begin{equation}
		{\mathbb{L}_{scpan}} =  {\mathbb{L}_{scl}} +  {\mathbb{L}_{org}} +  {\mathbb{L}_{coor}}
	\end{equation}
\end{small}
\begin{small}
	\begin{equation}
	{\mathbb{L}_{scl}} = \sum\limits_{n = 1}^N { {loss\left( {D\left( {{H^{n,\alpha }},B_c^{n,\alpha }} \right),\left( {{H^{n,\beta }},B_c^{n,\beta }} \right)} \right)}} 
	\end{equation}
\end{small}
\begin{small}
	\begin{equation}
	{\mathbb{L}_{org}} = \sum\limits_{n = 1}^N {\sum\limits_{k \in \{ \alpha ,\beta \} } {loss\left( {\left( {{H^{n,k}},B_{}^{n,k}} \right),\left( {{{\hat H}^{n,k}},\hat B_{}^{n,k}} \right)} \right)} }
	\end{equation}
\end{small}
\begin{small}
	\begin{equation}
		{\mathbb{L}_{coor}} = \left\| {\left( {\bar u,\bar v} \right) - \left( {{\hat u },{\hat v }} \right)} \right\|_2^2
	\end{equation}
\end{small}

\textbf{Original loss.} Generally, the Mean Square Error (MSE) loss is selected for optimizing models \cite{Yang2017StackedHN, Dong2018StyleAN, Bulat2018SuperFANIF, Chen2018FSRNetEL, Ma2020DeepFS} to generate landmark heatmaps. However, the MSE loss treats each pixel in the heatmap equally, which easily leads to blurred heatmaps and reduces the detection accuracy. Fortunately, MMDN \cite{Wan2021RobustFL} and MMHN \cite{Wan2021RobustFA} have shown that the Jensen-Shannon divergence loss can pay more attention to the foreground area of heatmaps instead of treating the whole heatmap equally, thus accurately measuring the difference between two distributions. Hence, the Jensen-Shannon Divergence loss is also selected as the objective function to calculate the distribution differences between the generated heatmaps and the ground-truths. The Jensen-Shannon Divergence loss is expressed as follows:

\begin{small}
	\begin{equation}
		\begin{array}{l}
			JS\left( {{p_G}||{p_{\hat G}}} \right) = \frac{1}{2}KL\left( {{p_G}\left( {i,j} \right)||\frac{{{p_G}\left( {i,j} \right) + {p_{\hat G}}\left( {i,j} \right)}}{2}} \right)\\
			{\kern 1pt} {\kern 1pt} {\kern 1pt} {\kern 1pt} {\kern 1pt} {\kern 1pt} {\kern 1pt} {\kern 1pt} {\kern 1pt} {\kern 1pt} {\kern 1pt} {\kern 1pt} {\kern 1pt} {\kern 1pt} {\kern 1pt} {\kern 1pt} {\kern 1pt} {\kern 1pt} {\kern 1pt} {\kern 1pt} {\kern 1pt} {\kern 1pt} {\kern 1pt} {\kern 1pt} {\kern 1pt} {\kern 1pt} {\kern 1pt} {\kern 1pt} {\kern 1pt} {\kern 1pt} {\kern 1pt} {\kern 1pt} {\kern 1pt} {\kern 1pt} {\kern 1pt} {\kern 1pt} {\kern 1pt} {\kern 1pt} {\kern 1pt} {\kern 1pt} {\kern 1pt} {\kern 1pt} {\kern 1pt} {\kern 1pt} {\kern 1pt} {\kern 1pt} {\kern 1pt} {\kern 1pt} {\kern 1pt} {\kern 1pt} {\kern 1pt} {\kern 1pt} {\kern 1pt} {\kern 1pt} {\kern 1pt} {\kern 1pt} {\kern 1pt} {\kern 1pt} {\kern 1pt} {\kern 1pt} {\kern 1pt} {\kern 1pt} {\kern 1pt}  + \frac{1}{2}KL\left( {{p_{\hat G}}\left( {i,j} \right)||\frac{{{p_G}\left( {i,j} \right) + {p_{\hat G}}\left( {i,j} \right)}}{2}} \right)
		\end{array}
	\end{equation}
\end{small}where $i$ and $j$ denote the indexes of a pixel in the heatmap and $KL$ means the Kullback-Leibler divergence. $p_{G}$ and $p_{\hat G}$ denote the probability distributions of the generated and ground-truth heatmaps. Base on the above Jensen-Shannon Divergence loss, the original loss can be reformulated as follows:

\begin{small}
	\begin{equation}
\begin{array}{l}
	{\mathbb{L}_{org}} = \sum\limits_{n = 1}^N \biggl\{ \sum\limits_{k \in \left\{ {\alpha ,\beta } \right\}} \Bigl\{ JS\left( {{p_{SE\left( {B_{u,v}^{n,k}} \right)}}||{p_{SE\left( {\hat B_{u,v}^{n,k}} \right)}}} \right)\\
	+ \{ JS\left( {p_{{H^{n,k}}}^T||p_{{{\hat H}^{n,k}}}^T} \right) + JS\left( {p_{B_c^{n,k}}^T||p_{\hat B_c^{n,k}}^T} \right)\} \\
	+ \sum\limits_{t = 1}^{T - 1}  \bigl\{ JS\left( {p_{{H^{n,k}}}^t||p_{{{\hat H}^{n,k}}}^t} \right) + JS\left( {p_{B_c^{n,k}}^t||p_{\hat B_c^{n,k}}^t} \right) \bigr\} \Bigr\} \biggr\} 
\end{array}
	\end{equation}
\end{small}where $t$ denotes the stage, and $T=4$. $SE(B_{u,v})$ means cropping the corresponding area (i.e., a square region with edge length $2r+1$ center at the ground-truth landmark location) from $B_u$ and $B_v$. 

\textbf{Self-calibrated loss.} The proposed self-calibrated mechanism could provide intermediate supervision by optimizing the loss of paired images. We also use the Jensen-Shannon Divergence loss as the objective function, and the self-calibrated loss can be reformulated as follows:

\begin{small}
	\begin{equation}
\begin{array}{l}
	{\mathbb{L}_{scl}} = \sum\limits_{n = 1}^N \Biggl \{ \sum\limits_{t = 1}^T \biggl [JS\left( {p_{D\left( {{H^{n,\alpha }}} \right)}^t||p_{{H^{n,\beta }}}^t} \right)\\
	+ JS\left( {p_{D\left( {B_c^{n,\alpha }} \right)}^t||p_{D\left( {B_c^{n,\beta }} \right)}^t} \right)\biggr ]\Biggr \} 
\end{array}
	\end{equation}
\end{small}

That means in each SCPA model, the self-calibrated loss is introduced to provide intermediate supervision and produce more effective pose attention masks for learning more discriminative representations.

\textbf{Coordinate loss.} The loss between the predicted landmarks and the ground-truths can also be used to optimize the proposed SCPAN, which further help obtain more accurate landmarks. The coordinate loss can be formulated as follows:

\begin{small}
	\begin{equation}
	{\mathbb{L}_{coor}} = \sum\limits_{n = 1}^N {\sum\limits_{k \in \{ \alpha ,\beta \} } {\sum\limits_{\phi  = 1}^\Phi  {\left\| {\left( {\bar u_\phi ^{n,k},\bar v_\phi ^{n,k}} \right) - \left( {\hat u_\phi ^{n,k},\hat v_\phi ^{n,k}} \right)} \right\|_2^2} } }  
	\end{equation}
\end{small}where $\phi$ denote the landmark index. The coordinate loss is able to help better integrate the boundary and field constraints and produce more effective landmark heatmaps and BALI fields for detecting more accurate landmarks.

\textbf{Overall loss.} By combining $\mathbb{L}_{org}$, $\mathbb{L}_{scl}$ and $\mathbb{L}_{coor}$, we can obtain the overall loss, which can be formulated as follows:

\begin{small}
	\begin{equation}
		\begin{array}{l}{\mathbb{L}_{scpan}} =  {\mathbb{L}_{scl}} +  {\mathbb{L}_{org}} +  {\mathbb{L}_{coor}}\\
			= \sum\limits_{n = 1}^N \Biggl\{ \sum\limits_{k \in \left\{ {\alpha ,\beta } \right\}} \biggl\{ {\lambda _1} \bigl\{ JS\left( {p_{{H^{n,k}}}^T||p_{{{\hat H}^{n,k}}}^T} \right) + JS\left( {p_{B_c^{n,k}}^T||p_{\hat B_c^{n,k}}^T} \right)\bigr\} \\
			+ \eta \sum\limits_{t = 1}^{T - 1}  \bigl \{ JS\left( {p_{{H^{n,k}}}^t||p_{{{\hat H}^{n,k}}}^t} \right) + JS\left( {p_{B_c^{n,k}}^t||p_{\hat B_c^{n,k}}^t} \right) \bigl \} \\
			+ {\lambda _2}JS\left( {{p_{SE\left( {B_{u,v}^{n,k}} \right)}}||{p_{SE\left( {\hat B_{u,v}^{n,k}} \right)}}} \right)\\
			+ \gamma \sum\limits_{\phi  = 1}^\Phi  {\left\| {\left( {\bar u_\phi ^{n,k},\bar v_\phi ^{n,k}} \right) - \left( {\hat u_\phi ^{n,k},\hat v_\phi ^{n,k}} \right)} \right\|_2^2} \biggr\} \\
			+ \eta \sum\limits_{t = 1}^{T - 1} \biggl \{ JS\left( {p_{D\left( {{H^{n,\alpha }}} \right)}^t\left\| {p_{{H^{n,\beta }}}^t} \right.} \right) + JS\left( {p_{D\left( {B_c^{n,\alpha }} \right)}^t||p_{D\left( {B_c^{n,\beta }} \right)}^t} \right) \biggr \} \Biggr\} 
		\end{array}
	\end{equation}
\end{small}

With the above overall loss function, the original label information can be better used to improve the detection accuray.

\subsection{Improving SCPAN with Semi-supervised Learning}
SCPAN is able to detect more accurate landmarks for faces with large poses and heavy occlusions by incorporating the proposed BALI fields and SCPA model. However, its performance still depends on large-scale training samples, i.e., high-resolution face images and their landmark annotations. Although it is easy to collect high-resolution face images, annotating them is more expensive and tedious. The proposed self-calibrated mechanism can provide a self-learned objective function that enforces intermediate supervision by utilizing unlabeled face data. In this way, more facial prior knowledge can be learned to enhance the detection accuracy of the proposed SCPAN (i.e., semi-SCPAN). Suppose there are $N$ labeled face images and $M$ unlabeled ones, the objective function of semi-SCPAN can be formulated as follows:

\begin{small}
	\begin{equation}
	\begin{array}{l}
		{{\mathbb{L}'}_{scpan}} = {\mathbb{L}_{scpan}} + {{\mathbb{L}'}_{scl}} = {\mathbb{L}_{scpan}}\\
		+ \sum\limits_{m = 1}^M \sum\limits_{t = 1}^{T-1} \left( JS\left( {{p^t_{D\left( {{H^{m,\alpha }}} \right)}}\left\| {{p^t_{{H^{m,\beta }}}}} \right.} \right){\kern 1pt}  + JS\left( {{p^t_{D\left( {B_c^{m,\alpha }} \right)}}\left\| {{p^t_{B_c^{m,\beta }}}} \right.} \right) \right) 
	\end{array}
\end{equation}\end{small}where ${{\mathbb{L}'}_{scl}}$ denotes the self-calibrated loss corresponding to $M$ unlabeled images and ${{\mathbb{L}'}_{scpan}}$ represents the final objective function of semi-SCPAN that is designed by utilizing both labeled and unlabeled data. With the above new objective function, the SCPAN is able to use unlabeled face images to further boost the performance of facial landmark detection. We also present the main steps of the proposed Semi-Supervised SCPAN in \textbf{Alogrithm I}.
\section{Experiments}
In this section, we firstly introduce the evaluation settings including the datasets and methods for comparison. Then, we compare our algorithm with the state-of-the-art facial landmark detection methods on challenging benchmark datasets such as 300W \cite{Sagonas2016300FI}, Menpo 2D \cite{Deng2018TheMB}, COFW \cite{Burgosartizzu2013Robust}, AFLW \cite{Zhu2016UnconstrainedFA}, WFLW \cite{Wu2018LookAB} and 300VW \cite{Shen2015TheFF}.
\subsection{Datasets and Implementation details}

\textbf{300W }(68 landmarks): The training set of 300W is composed of the training set of AFW \cite{Belhumeur2011LocalizingPO}, LFPW \cite{Zhu2012FaceDP} and Helen \cite{le2012interactive}, which contains 3148 face pictures. The testing set of 300W includes IBUG and the testing set of LFPW and Helen, which can be further divided as follows: 1) Challenging subset (i.e., IBUG dataset \cite{Sagonas2016300FI}). It contains 135 more general ``in the wild" images, and experiments for this dataset are more challenging. 2) Common subset (554 images, of which 224 images are from LFPW test set and 330 images from Helen test set). 3) Fullset (689 images, composed of the challenging subset and common subset). Moreover, following LUVLI \cite{Kumar2020LUVLiFA} and KDN \cite{Chen2018KernelDN}, we perform cross dataset evaluation on 300W dataset, i.e., We first train the SCPAN on 300W-LP dataset \cite{Zhu2016FaceAA}, and then fine-tune on the trainset (3837 samples). We evaluate the SCPAN on 300W testing set, which contains 600 images.

\textbf{Menpo 2D }(68 landmarks): It consists of images from AFLW and FDDB \cite{Jain2010FDDBAB}, which are re-annotated following 68 landmark annotation scheme. It has two subsets, frontal faces (6679 samples which have 68 annotations) and profile faces (300 samples which have 39 landmark annotations). We use the frontal set for cross dataset evaluation.

\textbf{COFW }(68 landmarks): It is another very challenging dataset on occlusion issues which is published by Burgos-Artizzu et al. \cite{Burgosartizzu2013Robust}. It contains 1345 training images of which 845 images are from LFPW and the others are with heavy occlusions. The testing set includes 507 face images with large variations on the head pose, facial expression and occlusion. We use the testing set for cross dataset evaluation.

\textbf{AFLW }(19 landmarks): It contains 25993 face images which has extremely large variations of jaw angles ranging from ${\rm{ - }}{120^ \circ }$ to ${\rm{ + }}{120^ \circ }$ and pitch angles ranging from ${\rm{ - }}{90^ \circ }$ to ${\rm{ + }}{90^ \circ }$. Moreover, it also contains very complicated occlusions. AFLW-full selects 24386 images from the whole AFLW dataset and further divides them into two parts: 20000 for training and 4386 for testing. Moreover, 1165 images (i.e., AFLW-frontal) are selected from AFLW-full testing set to evaluate the alignment algorithm on frontal faces.

\linespread{1.3}
\begin{table}
	\small
	\begin{center}
		\begin{tabular}{p{10cm}}
			\noindent\rule[0.25\baselineskip]{28em}{0.25pt}
			
			\hangafter=1
			\setlength{\hangindent}{8em}{\bfseries{Algorithm I}}:Semi-Supervised SCPAN
			
			\noindent\rule[0.25\baselineskip]{28em}{0.25pt}
			
			{\bfseries{Input}}: Training set $\{ {I^n}\} _{n = 1}^N$.
				
			{\bfseries{Output}}: Model $SCPAN$, BALI field $H$ and $B$, landmarks $\left( {\bar u,\bar v} \right)$.
			
			1: Uses the original face image ${I^{n,\alpha }}$ to generate its disturbance ${I^{n,\beta }}$.
			
			2: \textbf{IF} $I^n$ has landmark annotations \textbf{THEN}
			
			\hspace{2.5em}2.1 Generates the corresponding BALI field: 
			
			\hspace{2.5em} $\left( {{{\hat H}^{n,\alpha }},{{\hat B}^{n,\alpha }}} \right)$ and $\left( {{{\hat H}^{n,\beta }},{{\hat B}^{n,\beta }}} \right)$.
			
			\hspace{1.1em}\textbf{END IF} 
			
			3: \textbf{FOR epoch=1 to end\_epoch:}  
			
			\hspace{2.5em}3.1 Forward propagation of SCPAN model: 
			
			\hspace{2.5em} $\left( {{H^{n,k}},{B^{n,k}}} \right) = SCPAN({I^{n,k}}),{\rm{  }}k \in \left( {\alpha ,\beta } \right)$.
			
			\hspace{2.5em}3.2 Computes the loss: ${\mathbb{L}'}_{scpan} = {\mathbb{L}_{scpan}} + {{\mathbb{L}'}_{scl}}$.
			
			\hspace{2.5em}3.3 Calculates coarse landmarks: 
			
			\hspace{2.5em} $\left( {i',j'} \right) = \arg \mathop {\max }\limits_{i,j} \left( H \right)$.
			
			\hspace{2.5em}3.4 Calculates final landmarks: 
			
			\hspace{2.5em} $\left( {\bar u,\bar v} \right) = \frac{{\sum\nolimits_{i,j \in H'} {{H^{ij}} \times \left( {\left( {B_u^{ij},B_v^{ij}} \right) + \left( {i,j} \right)} \right)} }}{{\sum\nolimits_{i,j \in H'} {{H^{ij}}} }}$,
			
			\hspace{2.5em} $H' = \left\{ {\left( {i,j} \right)|i \in \left[ {i' - r,i' + r} \right],j \in \left[ {j' - r,j' + r} \right]} \right\}$.
			
			\hspace{2.5em}3.5 Updates the $SCPAN$ model.
			
			\hspace{1.1em}\textbf{END FOR} 
			
			\noindent\rule[0.25\baselineskip]{28em}{0.25pt}
		\end{tabular}
	\end{center}
	\label{tabalgorithm}
\end{table}

\textbf{WFLW }(98 landmarks): It has 98 landmark annotations and images in WFLW are collected from more complicated scenarios. The training set contains 7500 images and its testing set includes 2500 images. WFLW also has other attribute annotations, including occlusion, pose, makeup, lighting, blur, and expression, which can help more comprehensively evaluate existing alignment algorithms.

\textbf{300VW }(68 landmarks): Following Shen et al. \cite{Shen2015TheFF}, we use 50 videos to train our SCPAN and test it on the rest 64 videos. The testing set is divided into three parts: well-lit (\textbf{Scenario1}, various head poses, occlusions such as glasses and beards), mild unconstrained (\textbf{Scenario2}, different illuminations, dark rooms, overexposed shots and arbitrary expressions) and challenging (\textbf{Scenario3}, illumination conditions, occlusions, make-ups, expressions and head poses) according to the difficulties.

\textbf{Evaluation Metrics.} Facial landmark detection results are envaluated with Normalized Mean Error ($\rm NME_{box}$ \cite{Bulat2017HowFA, Zafeiriou2017TheMF}, $\rm NME_{diag}$ \cite{Sun2019HighResolutionRF, Wu2018LookAB} and $\rm NME_{io}$ \cite{Sagonas2016300FI, Tang2020TowardsEU}), Area Under the Curve (AUC) \cite{Kumar2020LUVLiFA, Wan2021RobustFA} and Failure Rate (FR) \cite{Chen2019FaceAW, Tang2020TowardsEU}. The $\rm{NME}$ is defined as follows:

\begin{small}
	\begin{equation}
	{\rm{NME}} = \frac{1}{\Phi }\sum\limits_{\phi  = 1}^\Phi  {\frac{{{{\left\| {\left( {{{\bar u}_\phi },{{\bar v}_\phi }} \right) - \left( {{{\hat u}_\phi },{{\hat v}_\phi }} \right)} \right\|}_2}}}{d}}
\end{equation}\end{small}where $\phi$ denotes the landmark index, $(\bar u, \bar v)$ and $(\hat u, \hat v)$ represent the predicted and ground-truth landmark coordinates, respectively. $d$ denotes the normalization term, which can be set to the interpupil distances, the interocular distance, the distance between the outer corners of the two eyes, the geometric mean of the width and height of the ground-truth bounding box and the diagonal of the tight bounding box for $\rm NME_{ip}$, $\rm NME_{io}$, $\rm NME_{box}$ and $\rm NME_{diag}$, respectively. To compute AUC, we firstly plot the cumulative distribution of the fraction of test images whose NME (\%) is less than or equal to the value on the horizontal axis, then the AUC is computed as the area under that curve. FR means the percentage of images in the test set whose NME is larger than a certain threshold.

\textbf{Implementation Details.} In our experiments, all the training and testing images are cropped and resized to 256x256 according to the provided bounding boxes. To generate the disturbance, we use the spatial transformation and texture disturbance operations. Specifically, the spatial transformation contains rotation ($-60^\circ, +60^\circ$), scaling ($0.5, 1$), mirror flip and their combinations. The texture disturbance includes occlusion, blurring and noises. The occlusion disturbance is achieved by using two types of occlusions. The first one is occlusion by black, i.e., the occluded area is covered by black, and the second one is occlusion by part of the original face image. The blurring disturbance means downsampling high-resolution images ($256\times256$) into low-resolution ($128\times128$, $64\times64$, $32\times32$, $16\times16$ ) ones with bicubic degradation. We use the Stacked Hourglass Network \cite{Yang2017StackedHN} as our backbone to construct the proposed Self-Calibrated Pose Attention Network, and the spatial size of the output heatmap and field is $128\times128$. $\lambda_1$, $\lambda_2$, $\gamma$ and $\eta$ are set to 1, 16, 40 and 4, respectively. $11 \times 11$ and $7\times7$ field regions are used in the training and testing phases, respectively. The training of SCPAN takes 200000 iterations and the staircase function is used to set the learning rate. The initial learning rate is $2.5 \times {10^{{\rm{ - 4}}}}$ and then it is divided by 5, 2 and 2 at iteration 10000, 40000 and 100000, respectively. The SCPAN is trained with Pytorch on 8 Nvidia Tesla V100 GPUs.

\textbf{Experiment Settings.} To evaluate the effectiveness of each module proposed in this paper, we firstly use the Stacked Hourglass Network (SHN) \cite{Yang2017StackedHN} as the baseline, and then we separately construct SHN+BALI and SHN+SCPA by combining SHN with the proposed Boundary-Aware Landmark Intensity (BALI) fields and Self-Calibrated Pose Attention (SCPA) model, respectively. Moreover, we combine SHN with both BALI fields and SCPA model to construct SCPAN (i.e., SHN+BALI+SCPA). Then, as SCPAN can be further boosted by utilizing unlabeled face images (in this paper, we use the CelebA \cite{Liu2015DeepLF} dataset) and the boosted SCPAN is denoted by semi-SCPAN. For CelebA, we use 169854 images for training. To be specific, we firstly use OpenFace \cite{Baltrusaitis2018OpenFace2F} to detect 68 landmarks and then obtain the corresponding input images with cutting and scaling operations according to the detected landmarks. For other state-of-the-art methods \cite{Kumar2018Disentangling3P, Dong2018StyleAN, Qian2019AggregationVS, Dong2019TeacherSS, Dapogny2019DeCaFADC, Kumar2020LUVLiFA, Wan2020RobustFL}, we either use the original codes released by the authors or restore the experiment, and the results both have achieved the expected effects in the corresponding papers. The detailed comparison is shown below.
\linespread{1}
\begin{table}
	\caption{$\rm NME_{io}$ and $\rm NME_{ip}$ comparisions on 300W dataset. (\% omitted)}
	\footnotesize
	\begin{center}
		\begin{tabular}{p{3cm}|p{1.2cm}p{1.2cm}p{0.8cm}}
			\hline
			Method  & 
			\begin{tabular}[c]{@{}c@{}}Common\\ Subset\end{tabular} & \begin{tabular}[c]{@{}c@{}}Challenging\\ Subset\end{tabular} & Fullset \\ \hline   
			$\rm NME_{io}$ comparisions
			\\ \hline                                                  
			PCD-CNN{$\rm _{CVPR18}$}[55]& 3.67 & 7.62      & 4.44    \\
			SAN{$\rm _{CVPR18}$}[30]           & 3.34    & 6.60      & 3.98    \\
			AVS{$\rm _{ICCV19}$}[56] & 3.21    & 6.49      & 3.86    \\
			LAB{$\rm _{CVPR18}$}[14]            & 2.98    & 5.19      & 3.49    \\
			Techer{$\rm _{ICCV19}$}[35]  & 2.91    & 5.91      & 3.49    \\
			DU-Net{$\rm _{ECCV18}$}[59]    & 2.90    & 5.15      & 3.35    \\
			DeCaFa{$\rm _{ICCV19}$}[57] & 2.93    & 5.26      & 3.39    \\
			HR-Net{$\rm _{19'}$}[50]               & 2.87    & 5.15      & 3.32    \\
			HG-HSLE{$\rm _{ICCV19}$}[60] & 2.85    & 5.03      & 3.28    \\
			AWing{$\rm _{ICCV19}$}[61]     & 2.72  & 4.52   & 3.07 \\
			LUVLi{$\rm _{CVPR20}$}[12]       & 2.76  & 5.16    & 3.23  \\
			CCDN{$\rm _{NN21}$}[58]       & 2.75  & 4.43    & 3.08  \\
			 \hline
			SHN  & 3.11  & 6.23    & 3.72  \\ 
			\textbf{SHN+BALI} & \textbf{2.78}  & \textbf{5.01}    & \textbf{3.21}  \\ 
			\textbf{SHN+SCPA} & \textbf{2.59}  & \textbf{4.68}    & \textbf{3.00}  \\ 
			\textbf{SCPAN} & \textbf{2.46}  & \textbf{4.43}    & \textbf{2.85}  \\ 
			\textbf{semi-SCPAN} & \textbf{2.38} & \textbf{4.31}    & \textbf{2.76}  \\ \hline
			$\rm NME_{ip}$ comparisions
			\\ \hline
			Honari et al.{$\rm _{CVPR18}$}[33] &4.20 &7.78 &4.90 \\
			SBR{$\rm _{CVPR18}$}[34] &3.28 &7.58 &4.10 \\
			TS{$\rm _{CVPR18}$}[35] &3.17 &6.41 &3.78 \\
			Liu et al.{$\rm _{CVPR19}$}[31] &	3.45&	6.38&	4.02\\
			ODN{$\rm _{CVPR19}$}[29]	&3.56	&6.67&	4.17\\
			STKI{$\rm _{ACM MM20}$}[7]	&3.36	&7.39&	4.16\\
			MMDN{$\rm _{TNNLS21}$}[15]	&3.17	&6.08&	3.74\\\hline
			\textbf{semi-SCPAN} & \textbf{2.92} & \textbf{5.96}    & \textbf{3.52}  \\ \hline
		\end{tabular}
	\end{center}
	\label{tab300w}
	\vspace{-1em}
\end{table}
\subsection{Evaluation under Normal Circumstances}
For benchmark datasets such as 300W, Menpo 2D, COFW, AFLW and WFLW, faces in 300W common subset, 300W full set, Menpo 2D and AFLW-frontal dataset are closer to neutral faces and have smaller variations on the head pose, facial expression and occlusion. Hence, we evaluate the effectiveness of the proposed SCPAN method under normal circumstances with these four subsets. Table \ref{tab300w} includes the $\rm NME_{io}$ and $\rm NME_{ip}$ comparisons of state-of-the-art face alignment methods on 300W common subset and 300W full set. Moreover, Table \ref{tabpretrain} shows the {$\rm  NME_{box}$ and $\rm AUC^7_{box}$ comparisons on 300W testing set and Menpo 2D (for cross dataset evaluation). As shown in Tables  \ref{tab300w} and \ref{tabpretrain}, the proposed SCPAN outperforms state-of-the-art methods \cite{Dong2018StyleAN, Dong2019TeacherSS, Dapogny2019DeCaFADC, Sun2019HighResolutionRF, Wang2019AdaptiveWL, Kumar2020LUVLiFA} on 300W and Menpo 2D datasets. At the same time, the SCPAN also achieves the best score on AFLW-frontal dataset (as shown in Table \ref{tabaflw}). These results indicate that the SCPAN can improve the detection accuracy under normal circumstances, mainly because 1) the proposed BALI fields can introduce both boundary and field constraints to enhance facial shape constraints and achieve high-precision landmark detection. 2) the SCPA model can boost supervision of the network and selectively emphasize important features and suppress less useful ones to learn more representative and discriminative features by introducing the self-calibrated mechanism and the pose attention mask. 3) by integrating the SCPA model and BALI fields into a Self-Calibrated Pose Attention Network (SCPAN), more accurate and effective landmark heatmaps and BALI fields can be generated for achieving robust and precise facial landmark detection.
\begin{table}
	\caption{$\rm NME_{box}$ and $\rm AUC^7_{box}$ comparisions on 300W Common subset, Menpo 2D and COFW-68 datasets. (- not counted, \% omitted) [* = pretrained on 300W-LP-2D \cite{Zhu2016FaceAA}]}
	\begin{center}
		\footnotesize
		\begin{tabular}{p{2.6cm}|p{0.5cm}p{0.5cm}p{0.6cm}|p{0.5cm}p{0.5cm}p{0.6cm}}
			\hline
			\multirow{2}{*}{Method} & \multicolumn{3}{|c}{$\rm NME_{box}$} & \multicolumn{3}{|c}{$\rm AUC^7_{box}$}  \\
			\cline{2-7}
			& 300W &Menpo  & COFW & 300W &Menpo & COFW \\
			\cline{2-7}	
			\cline{0-0}
			SAN*{$\rm _{CVPR18}$}[30] & 2.86 & 2.95 & 3.50 & 59.7 &61.9 & 51.9  \\	
			2D-FAN*{$\rm _{ICCV17}$}[48] & 2.32 &2.16 & 2.95 & 66.5 &69.0 & 57.5  \\
			KDN[45] & 2.49 &2.26 & - & 67.3 &68.2 & -  \\
			Softlabel*{$\rm _{ICCV19}$}[52] & 2.32 &2.27 & 2.92 & 66.6 &67.4 & 57.9  \\
			KDN*{$\rm _{ICCV19}$}[52] & 2.21 &2.01 & 2.73 & 68.3 &71.1 & 60.1  \\
			LUVLi{$\rm _{CVPR20}$}[12] & 2.24 &2.18 & 2.75 & 68.3 &70.1 & 60.8  \\	
			LUVLi*{$\rm _{CVPR20}$}[12] & 2.10 &2.04 & 2.57 & 70.2 &71.9 & 63.4  \\ \hline
			\textbf{SCPAN} & \textbf{2.01} &\textbf{1.93} & \textbf{2.47} & \textbf{71.8} &\textbf{72.8} & \textbf{65.1}  \\
			\textbf{SCPAN*} & \textbf{1.95} &\textbf{1.88} & \textbf{2.38} & \textbf{72.5} &\textbf{73.1} & \textbf{65.7}  \\ \hline
		\end{tabular}
	\end{center}
	\label{tabpretrain}
\end{table}
\begin{table}
	\caption{$\rm NME$ and $\rm AUC$ comparisions on AFLW dataset. (- not counted, \% omitted)}
	\begin{center}
		\footnotesize
		\begin{tabular}{p{2.3cm}|p{0.8cm}p{0.9cm}|p{1cm}|p{1cm}}
			\hline
			\multirow{2}{*}{Method} & \multicolumn{2}{|c|}{$\rm NME_{diag}$} &{$\rm NME_{box}$} & {$\rm AUC^7_{box}$} \\
			\cline{2-5}
			& Full & Frontal & Full & Full \\
			\cline{2-5}	
			\cline{0-0}
			CCL{$\rm _{CVPR16}$}[21] & 2.72 & 2.17 & - & -  \\
			LLL{$\rm _{ICCV19}$}[62] & 1.97 & - & - & -  \\
			SAN{$\rm _{CVPR18}$}[30] & 1.91 & 1.85 & 4.04 & 54.0  \\
			DSRN{$\rm _{CVPR18}$}[63] & 1.86 & - & - & -  \\	
			LAB{$\rm _{CVPR18}$}[14] & 1.85 & 1.62 & - & -  \\
			HR-Net{$\rm _{19'}$}[50] & 1.57 & 1.46 & - & -  \\
			Wing{$\rm _{CVPR18}$}[64] & - & - & 3.56 & 53.5  \\ 
			KDN[45] & - & - & 2.80 & 60.3  \\ 
			LUVLi{$\rm _{CVPR20}$}[12] & 1.39 & 1.19 & 2.28 & 68.0  \\ 
			MHHN{$\rm _{TIP21}$}[13] & 1.38 & 1.19 & - & -  \\ 
			\hline
			SHN & 2.46 & 1.92 &3.67 &56.4  \\ 
			\textbf{SHN+BALI} & \textbf{1.84} & \textbf{1.46} & \textbf{2.32} & \textbf{66.1}  \\ 
			\textbf{SHN+SCPA} & \textbf{1.62} & \textbf{1.37} & \textbf{2.21} & \textbf{67.2}  \\
			\textbf{SCPAN} & \textbf{1.31} & \textbf{1.10} & \textbf{2.05} & \textbf{69.8}  \\
			\textbf{semi-SCPAN} & \textbf{1.23} & \textbf{1.05} & \textbf{2.01} & \textbf{70.7}  \\ \hline
		\end{tabular}
	\end{center}
	\label{tabaflw}
\end{table}
\begin{table}
	\caption{$\rm NME_{ip}$ and $\rm FR_{ip}^{10}$ comparisons on COFW dataset. (- not counted, \% omitted)}
	\begin{center}
		\footnotesize
		\begin{tabular}{p{4.6cm}p{1cm}p{1.0cm}}
			\hline
			Method & $\rm NME_{ip}$  & $\rm FR_{ip}^{10}$ \\
			\hline
			DRDA{$\rm _{CVPR16}$}[65] &	6.46&	6.00\\
			RAR{$\rm _{ECCV16}$}[66] &	6.03&	4.14\\
			DAC-CSR{$\rm _{CVPR17}$}[67] &	6.03&	4.73\\
			CAM{$\rm _{19'}$}[68] &	5.95&	3.94\\
			PCD-CNN{$\rm _{CVPR18}$}[55] & 5.77 &3.73\\
			Wing{$\rm _{CVPR18}$}[64] &	5.44&	3.75\\
			LAB{$\rm _{CVPR18}$}[14] &	5.58&	2.76\\
			AWing{$\rm _{ICCV19}$}[6] &4.94 &	0.99\\
			ODN{$\rm _{CVPR19}$}[29] &5.30 &	-\\
			MHHN{$\rm _{TIP21}$}[13] &4.95 &1.78 \\
			\hline
				SHN &	6.21&	5.52\\
			\textbf{SHN+BALI} &	\textbf{5.52}&	\textbf{3.16} \\
			\textbf{SHN+SCPA} &	\textbf{5.21}&	\textbf{2.17} \\
			\textbf{SCPAN} &	\textbf{4.93}&	\textbf{1.78} \\
			\textbf{semi-SCPAN} &	\textbf{4.83}&	\textbf{1.58} \\
			\hline
		\end{tabular}
	\end{center}
	\label{tabcofw}
\end{table}
\begin{table*}
	\caption{$\rm NME_{io}$ comparisons on WFLW dataset. (\% omitted).}
	\begin{center}
		\begin{tabular}{p{3cm}p{1.3cm}p{1.3cm}p{1.5cm}p{1.7cm}p{1.5cm}p{1.3cm}p{1.2cm}}
			\hline
			Method  &
			\begin{tabular}[l]{@{}l@{}}Testset\\\end{tabular} & \begin{tabular}[l]{@{}l@{}}Pose\\ Subset\end{tabular} &
			\begin{tabular}[c]{@{}l@{}}Expression\\ Subset\end{tabular} &
			\begin{tabular}[c]{@{}l@{}}Illumination\\ Subset\end{tabular} &
			\begin{tabular}[c]{@{}l@{}}Make-Up\\ Subset\end{tabular} &
			\begin{tabular}[c]{@{}l@{}}Occlusion\\ Subset\end{tabular} &
			\begin{tabular}[c]{@{}l@{}}Blur\\ Subset\end{tabular}  \\ \hline
			CCFS{$\rm _{CVPR15}$}[69] &9.07 &21.36 &10.09 &8.30 &8.74 &11.76 &9.96 \\
			DVLN{$\rm _{CVPR17}$}[70] &6.08 &11.54 &6.78 &5.73 &5.98 &7.33 &6.88 \\
			LAB{$\rm _{CVPR18}$}[14] &5.27 &10.24 &5.51 &5.23 &5.15 &6.79 &6.32 \\
			Wing{$\rm _{CVPR18}$}[64] &5.11 &8.75 &5.36 &4.93 &5.41 &6.37 &5.81 \\
			MHHN{$\rm _{TIP21}$}[13] &4.77 &9.31 &4.79 &4.72 &4.59 &6.17 &5.82 \\
			MMDN{$\rm _{TNNLS21}$}[15]	&4.87 &8.15 &4.99 &4.61 &4.72 &6.17 &5.72 \\
			HRNet{$\rm _{19'}$}[50] &4.60 &7.86 &4.78 &4.57 &4.26 &5.42 & 5.36 \\
			AWing{$\rm _{ICCV19}$}[61] &4.36 &7.38 &4.58 &4.32 &4.27 &5.19 &4.96 \\
			LUVLi{$\rm _{CVPR20}$}[12] &4.37 &7.56 &4.77 &4.30 &4.33 &5.29 &4.94 \\
			\hline
			SHN &5.78 &9.47 &6.39 &5.83 &5.91 &7.07 &7.21 \\
			\textbf{SHN+BALI} &\textbf{4.73} &\textbf{7.67} &\textbf{4.96} &\textbf{4.73} &\textbf{4.56} &\textbf{5.67} &\textbf{5.47} \\
			\textbf{SHN+SCPA} &\textbf{4.57} &\textbf{7.43} &\textbf{4.77} &\textbf{4.52} &\textbf{4.38} &\textbf{5.44} &\textbf{5.12} \\
			\textbf{SCPAN} &\textbf{4.29} &\textbf{7.22} &\textbf{4.68} &\textbf{4.34} &\textbf{4.21} &\textbf{5.25} &\textbf{4.88}\\
			\textbf{semi-SCPAN} &\textbf{4.21} &\textbf{7.01} &\textbf{4.57} &\textbf{4.15} &\textbf{4.16} &\textbf{5.17} &\textbf{4.81}\\
			\hline
		\end{tabular}
	\end{center}
	\label{tabwflw}
\end{table*}
\subsection{Evaluation of Robustness against Occlusion}
Variations in occlusion and illumination are classic problems in face alignment task. The state-of-the-art face alignment methods still suffer from heavy occlusions and complicated illuminations. In this paper, we use COFW dataset, 300W challenging subset and WFLW dataset to evaluate the robustness of the proposed SCPAN against occlusion.

For 300W challenging subset, SCPAN can achieve 4.43\% $\rm NME_{io}$ as shown in Table \ref{tab300w}, which outperforms state-of-the-art face alignment methods \cite{Dong2018StyleAN, Dong2019TeacherSS, Dapogny2019DeCaFADC, Sun2019HighResolutionRF, Wang2019AdaptiveWL, Kumar2020LUVLiFA}. This indicates the method can effectively enhance the alignment robustness for faces with heavy occlusions.

For COFW dataset (cross dataset evaluation), the failure rate ($\rm FR_{ip}^{10}$) is defined by the percentage of test images with more than 10\% detection error which is normalized by interpupil distance. As illustrated in Table \ref{tabcofw}, the SCPAN can boost the $\rm NME_{ip}$ to 4.93\% and the failure rate to 1.78\%, which outperforms the state-of-the-art methods \cite{Wan2019FaceAB, Kumar2018Disentangling3P, Feng2017WingLF, Wu2018LookAB, Wang2019AdaptiveWL, Zhu2019RobustFL, Wan2021RobustFA}. Moreover, we also use the model trained on 300W training set to further test COFW testing set and the corresponding experimental results (i.e., $\rm NME_{box}$ and $\rm AUC^7_{box}$) are shown in Table \ref{tabpretrain}. These all suggest that the proposed BALI fields and SCPA model play an important role in boosting the ability to address the occlusion problems.

Since WFLW dataset is composed of the Illumination subset, the Make-Up Subset and the Occlusion Subset, which contain complicated occlusions that can be used to evaluate the robustness of the SCPAN against occlusion. From the experimental results shown in Table \ref{tabwflw}, we conclude that SCPAN is more robust to faces with complicated occlusions.

Hence, from the experimental results illustrated in Tables \ref{tab300w}, \ref{tabpretrain}, \ref{tabcofw} and \ref{tabwflw}, we can conclude that 1) by fusing boundary heatmaps and landmark intensity fields, the proposed BALI fields is able to better model the facial geometric information and context information for enhancing facial shape constraints. 2) the SPCA model can learn more representative and discriminative features by introducing the self-calibrated mechanism and pose attention mask when faces are corrupted by complicated occlusions. 3) the texture disturbance operation can effectively expand the original dataset and make full use of the original label information, which helps achieve more robust landmark detection for faces with heavy occlusions.
\subsection{Evaluation of Robustness against Large Poses}
Face with large poses is another great challenge for facial landmark detection. We conduct experiments on AFLW-full, 300W challenging subset and WFLW dataset to further evaluate the performance of the SCPAN for faces with large poses. Fig. \ref{landmarks} shows the landmark detection result comparison with the state-of-the-art methods and the ground-truths (i.e., GT) on 300W challenging subset. Tables \ref{tab300w}, \ref{tabaflw} and \ref{tabwflw} show the corresponding experimental results. To be specific, for AFLW-full dataset, the proposed SCPAN can achieve 1.31\% $\rm  NME_{diag}$, 2.05\% $\rm NME_{box}$ and 69.8\% $\rm AUC_{box}^7$, which exceeds state-of-the-art methods \cite{Robinson2019LaplaceLL, Dong2018StyleAN, Wu2018LookAB, Sun2019HighResolutionRF, Kumar2020LUVLiFA, Wan2021RobustFA}. For WFLW dataset, the $\rm NME_{io}$ on Testset, Pose Subset and Expression Subset all beat the other state-of-the-art methods \cite{wu2017leveraging, Wu2018LookAB, Wan2021RobustFA, Sun2019HighResolutionRF, Wang2019AdaptiveWL, Kumar2020LUVLiFA}. From the above experimental results, we can conclude that our method is more robust to faces with large poses, mainly because 1) the spatial transformation disturbance operations can effectively enrich the original dataset, i.e., produce many face images with large poses, thus the robustness of our method against large poses has been improved. 2) the SCPAN can better model the facial shape constraints by integrating the BALI fields and SCPA model, which improves the accuracy of landmark detection for faces with large poses.
\begin{figure}[t]
	\begin{center}
		\includegraphics[width=0.98\linewidth]{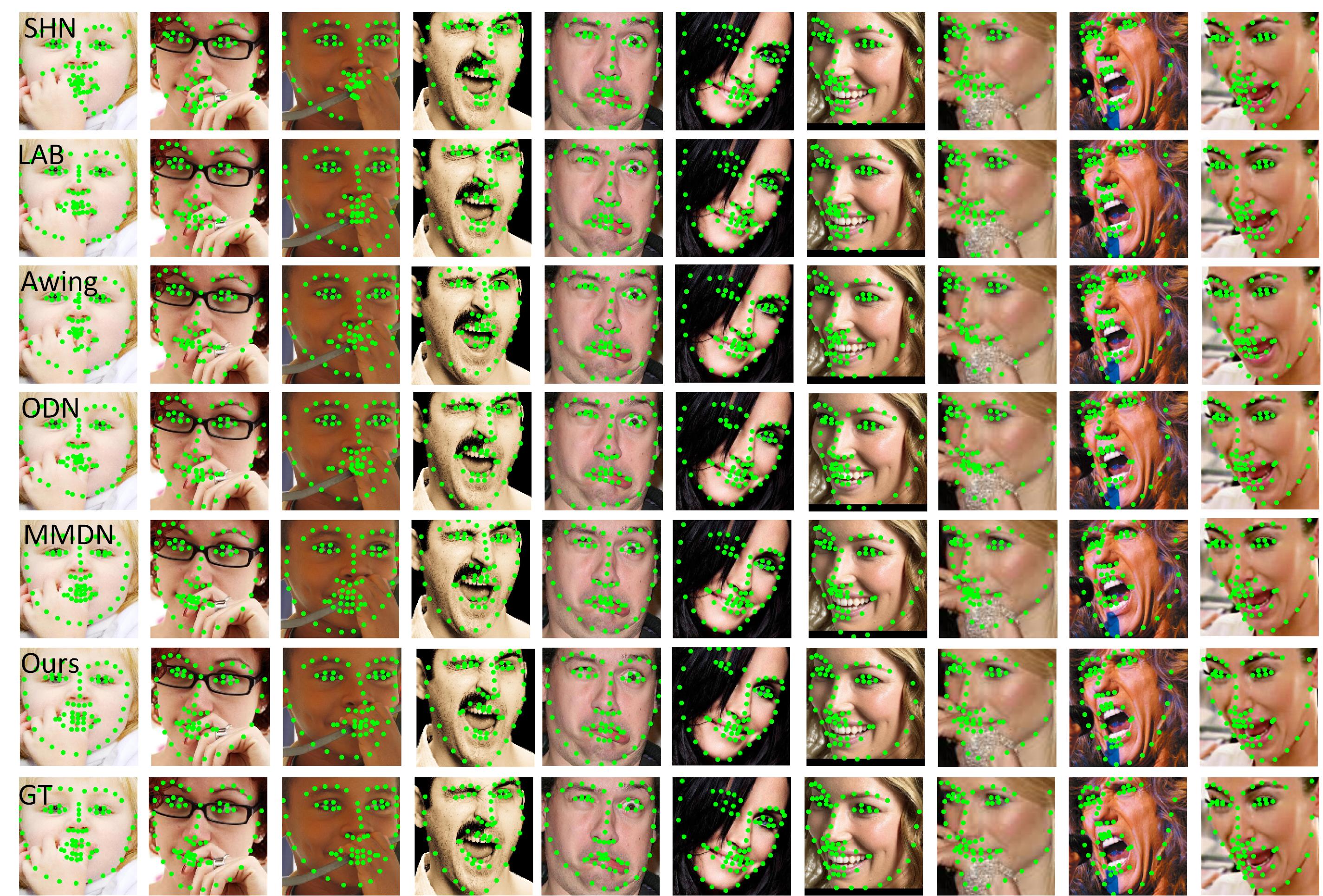}
	\end{center}
	\caption{Landmark detection result comparison with the state-of-the-art methods on 300W challenging subset. Our proposed method is more robust to faces under large poses and heavy occlusions.}
	\label{landmarks}
\end{figure}

\begin{table}
	\caption{$\rm NME_{ip}$ comparisons on 300VW dataset. (\% omitted)}
	\begin{center}
		\footnotesize
		\begin{tabular}{p{3.2cm}p{1.2cm}p{1.2cm}p{1.2cm}}
			\hline
			Method & Scenario1  & Scenario2 & Scenario3 \\
			\hline
			TSCN{$\rm _{NIPS14}$}[71]  &12.54  &7.25  &13.13 \\
			CCFS{$\rm _{CVPR15}$}[69]  &7.68 &6.42 &13.67 \\
			TCDCN{$\rm _{TPAMI16}$}[72]  &7.66 &6.77 &14.98 \\
			CCR{$\rm _{ECCV16}$}[73]  &7.26 &5.89 &15.74 \\
			iCCR{$\rm _{ECCV16}$}[73]  &6.71 &4.00 &12.75 \\
			MDM{$\rm _{CVPR16}$}[28]  &5.46 &4.59 &7.42 \\
			TSTN{$\rm _{TPAMI18}$}[74]  &5.36 &4.51 &12.84 \\
			FHR{$\rm _{AAAI19}$}[75]  &5.07 &4.34 &7.36 \\
			FHR+STA{$\rm _{AAAI19}$}[75]  &4.42 &4.18 &5.98 \\
			STKI{$\rm _{ACM MM20}$}[7]  &5.04 &4.57 &6.11 \\
			\hline
			\textbf{SCPAN} &	\textbf{4.49}&	\textbf{4.23} &	\textbf{5.87}\\
			\hline
		\end{tabular}
	\end{center}
	\label{tab300VW}
\end{table}

\subsection{Evaluation on Face Videos}
We evaluate our proposed SCPAN on 300VW dataset with state-of-the-art facial landmark detection methods \cite{zhu2015face, Zhang2016LearningDR, Liu2018TwoStreamTN, Tai2019TowardsHA, Zhu2020SpatialTemporalKI}. From the experimental results as shown in Table \ref{tab300VW}, we can find that (1) on both \textbf{Scenario1} and \textbf{Scenario2}, our SCPAN can achieve state-of-the-art accuracy. (2) on \textbf{Scenario3}, our SCPAN beats the best score. These indicate that our proposed SCPAN is more robust to faces under variations on illumination conditions, occlusions, make-ups, expressions and head poses. Moreover, we believe that our SCPAN structure can be further boosted by using a temporal modeling technique \cite{Tai2019TowardsHA, Zhu2020SpatialTemporalKI}.

\subsection{Evaluation of Semi-SCPAN}
Current facial landmark detection methods still suffer insufficient labeled training samples.  With a well-designed SCPA model, the proposed SCPAN is able to use unlabeled face images to boost the detection accuracy. To evaluate this, we use both labeled and unlabeled datasets to train SCPAN, i.e., 300W, AFLW, COFW and WFLW are separately mixed with CelebA dataset to train SCPAN (denoted as semi-SCPAN), the corresponding experimental results are shown in Table \ref{tab300w}, \ref{tabaflw}, \ref{tabcofw} and \ref{tabwflw}, respectively. From this we can see that semi-SCPAN outperforms the state-of-the-art methods \cite{wu2017leveraging, Dong2018StyleAN, Wu2018LookAB, Wan2021RobustFA, Sun2019HighResolutionRF, Wang2019AdaptiveWL, Kumar2020LUVLiFA} and other semi-supervised methods \cite{Honari2018ImprovingLL, Dong2018SupervisionbyRegistrationAU, Dong2019TeacherSS, Zhu2020SpatialTemporalKI}. This indicates that by introducing unlabeled face datasets, more prior knowledge can be learned to model facial shape constraints by semi-SCPAN for boosting the detection accuracy.
\subsection{Self Evaluations}
\textbf{Heatmap generation.} Almost heatmap regression-based facial landmark detection methods use heatmaps generated by a Gaussian distribution to regress and predict landmark coordinates. The closer to the landmark, the greater the belief value is. Moreover, the gradient of belief value is also changed. Therefore, by using the Gaussian distribution, neural networks can quickly and directionally reach the landmark. We also explore other two non-Gaussian distributions, i.e., Generalized Error Distribution (GED) \cite{Mohseni2014NonGaussianPM, Havyarimana2021TowardAI} and Student-t Distribution (StD) \cite{Liang2013IndependentVA} to generate landmark heatmaps and boundary heatmaps. Note that, when we set $df =  + \infty$ or $d=0$, the curves of GED and StD become the Standard Normal Distribution. $df$ means the degree of freedom of StD, $d$ denotes the shape parameter of GED. As shown in Table \ref{tabeffect}, we can find that using non-Gaussian distributions to generate heatmaps can achieve comparable or even better results than using a Gaussian distribution.

\begin{table}
	\caption{The effect ($\rm NME_{io}$ (\%)) of different heatmaps on 300W challenging subset.}
	\begin{center}
		\footnotesize
		\begin{tabular}{p{1.2cm}|p{0.6cm}|p{0.6cm}p{0.6cm}p{0.8cm}|p{0.6cm}p{0.6cm}}
			\hline
			\multirow{2}{*}{Distribution} & \multicolumn{1}{|c}{Gaussian} & \multicolumn{3}{|c}{GED} & \multicolumn{2}{|c}{StD} \\
			\cline{2-7}
			& - & d=0.2 & d=0.1 & d=-0.1 & df=1 & df=3 \\
			\cline{2-7}	
			\cline{0-0}
		   NME &4.43	&4.67	&4.39	&5.01	&4.42	&4.40  \\
		 \hline
		\end{tabular}
	\end{center}
	\label{tabeffect}
\end{table}

\textbf{Sensitivity analysis of parameters.} By combing $\mathbb{L}_{org}$, $\mathbb{L}_{scl}$ and $\mathbb{L}_{coor}$, we can obtain the overall loss. The losses corresponding to landmark heatmap and boundary heatmap should be given the same weight. However, the sizes of heatmaps in $t = T$ and $t = 1 \cdots T - 1$ are $128\times128$ and $64 \times 64$, respectively. Therefore, $\lambda_1$ and $\eta$ are set to 1 and 4, respectively. Besides, the loss of field is calculated with a smaller area, $\lambda_2$ should be given a larger weight (i.e., 16). When calculating $\mathbb{L}_{coor}$, we first normalize the landmark coordinates to $[0, 1]$, and then $\gamma$ is set to 40. We also conduct the corresponding experiments by using different $\lambda_1$,  $\lambda_2$, $\gamma$ and $\eta$ values on 300W challenging subsets. From the experimental results in Tabel \ref{tablepara}, we can find that $\lambda_1=1$, $\lambda_2=16$, $\gamma=40$ and $\eta=4$ are good choices to balance these four parts.

\begin{table}
	\caption{The effect ($\rm{NME_{io}}$(\%)) of different  $\lambda_1$,  $\lambda_2$, $\gamma$ and $\eta$ values on 300W challenging subset.}
	\begin{center}
		\footnotesize
		\begin{tabular}{p{1.2cm}p{1.2cm}p{1.2cm}p{1.2cm}p{1.20cm}}
			\hline
			$\lambda_1$ & $\eta$  & $\lambda_2$ & $\gamma$ & NME \\
			\hline
			1	 &0	 &0	 	 &0	 	&5.96\\
			1	 &4	 &0	 	 &0	 	&5.69\\
			1	 &4	 &8	 	 &0	 	&4.96\\
			1	 &4	 &16	 &0	 	&4.78\\
			1	 &4	 &32	 &0	 	&4.84\\
			1	 &4	 &16	 &20	&4.54\\
			1	 &4	 &16	 &40	&4.43\\
			1	 &4	 &16	 &60	&5.01\\
			\hline
		\end{tabular}
	\end{center}
	\label{tablepara}
\end{table}

\textbf{Time and memory analysis.} The proposed SCPAN contains three SCPA models and an Hourglass Network Unit (HNU). Compared to the HNU, our SCPA model increase both the parameter and computational costs. For the baseline i.e., four stacked SHN, its model size and parameters are 16.05MB and 184MB, respectively. While Our SCPAN takes 16.99MB parameters and its model size is 195MB. Besides, the reference speed of our SCPAN can achieve 40FPS in a single Telsa V100 GPU while the SHN achieves 100FPS. If we use multiple GPUs, the reference will speed up. Among all, compared to SHN, our SCPAN introduces some additional parameters which are insignificant compared to 16GB memory on a tesla v100, and the impact of the consumed computation costs will be decreased with the rapid development of the hardware. To reduce the computational costs or employ lightweight backbones, we also conduct the experiment by reducing the number of SCPA models, i.e., one SCPA model and one HNU are used to construct SCPAN. Its parameters and model size will be reduced to 8.73MB and 100MB, respectively. But the $\rm{NME_{io}}$ on 300W challenging dataset will only rise from 4.43\% to 4.61\%, which still achieves state-of-the-art accuracy.

\textbf{Visualization of the proposed BALI fields.} Since the proposed SCPAN can produce more effective landmark heatmaps and BALI fields and help achieve more accurate facial landmark detection, we visualize the produced landmark heatmaps and BALI fields in Fig. \ref{ger_field}. As shown there, the first three rows visualize the landmark heatmaps, boundary heatmaps, BALI fields, the x-offset of the BALI fields and the y-offset of the BALI fields for faces under large poses, and the last three rows show those for faces with complicated occlusions. From Fig. \ref{ger_field}, we can see that the SCPAN is able to generate effective and accurate landmark heatmaps and BALI fields for faces in challenging scenarios. Therefore, SCPAN achieves state-of-the-art performance on challenging benchmark datasets.
\begin{figure}[t]
	\begin{center}
		\includegraphics[width=0.92\linewidth]{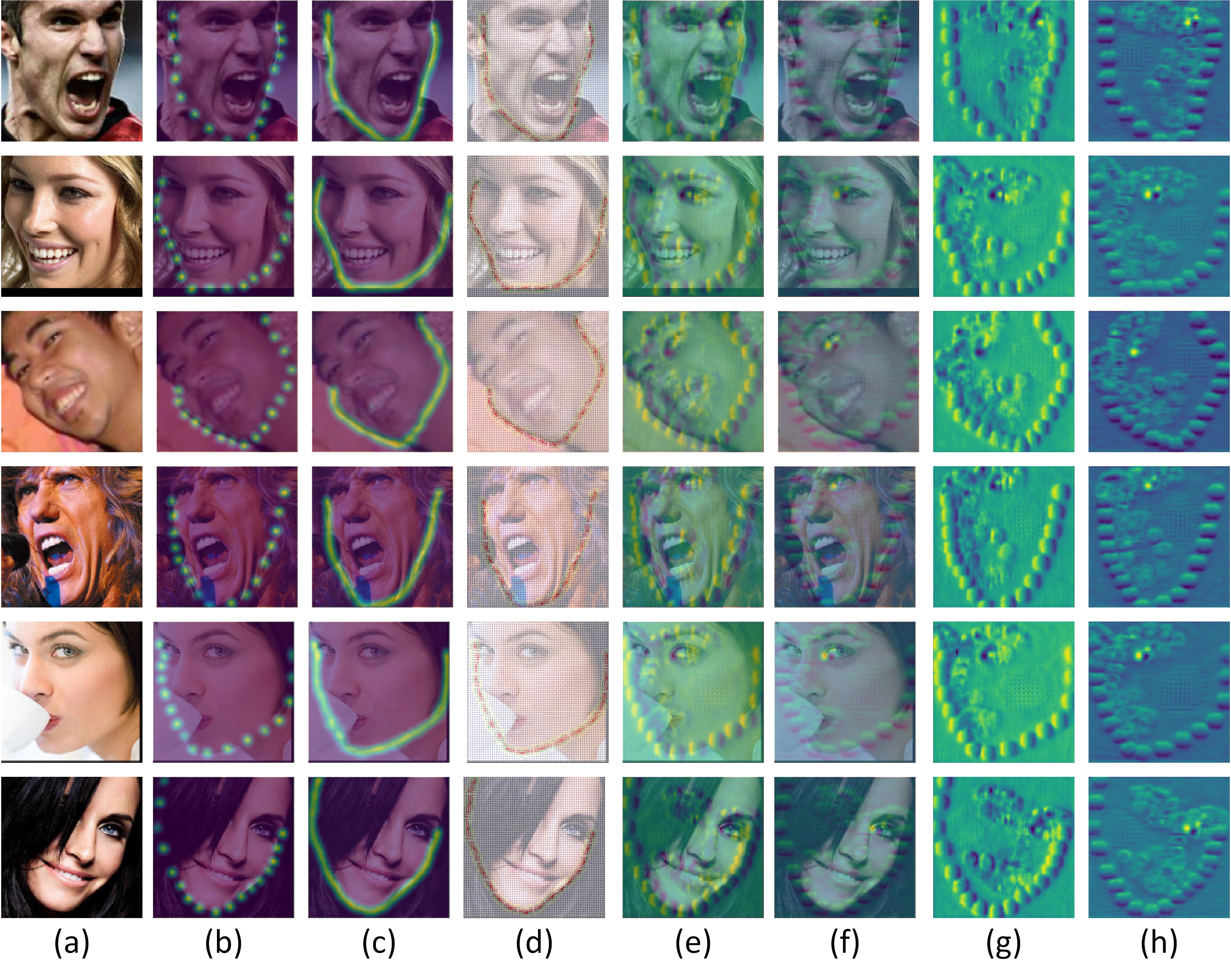}
	\end{center}
	\caption{Visualization of the proposed BALI fields. (a) The original image. (b) landmark heatmaps. (c) boundary heatmaps. (d) BALI fields. (e) the fusion of the x-offset of BALI field and the original image. (f) the fusion of the y-offset of BALI field and the original image. (g)the x-offset of BALI field. (h) the y-offset of BALI field. The SCPAN can produce more accurate and effective landmark heatmaps and BALI fields, thus achieving state-of-the-art performance.}
	\label{ger_field}
\end{figure}
\subsection{Ablation Studies}
The proposed Self-Calibrated Pose Attention Network (SCPAN) contains two pivotal components, namely, the BALI fields and the SCPA model. Moreover, SCPAN (denoted as semi-SCPAN) can be further boosted by using unlabeled images. Therefore, the ablation studies are conducted as follows:

(1) The proposed BALI fields can introduce both boundary and field constraints to the predicted landmarks. To evaluate this, we conduct the experiment by combining SHN \cite{Yang2017StackedHN} with the proposed BALI fields (denoted as SHN+BALI) on challenging benchmark datasets including 300W, AFLW, COFW and WFLW. From the experimental results as shown in Tables \ref{tab300w}, \ref{tabaflw}, \ref{tabcofw} and \ref{tabwflw}, we can see that SHN+BALI outperforms SHN, which indicates that the proposed BALI fields can better model the facial shape constraints and help detect more accurate landmarks.

(2) The proposed SCPA model is able to learn more representative and discriminative features for producing more effective landmark heatmaps, hence, we conduct the experiment by combining SHN and SCPA model (denoted as SHN+SCPA). The experimental results in Tables \ref{tab300w}, \ref{tabaflw}, \ref{tabcofw} and \ref{tabwflw} shows that SHN+SCPA outperforms SHN, which demonstrates that the proposed SCPA model is able to generate more effective landmark heatmaps and help detect more accurate landmarks by introducing a self-calibrated mechanism and a pose attention mask.

(3) The proposed SCPAN is constructed by combining the SHN \cite{Yang2017StackedHN} with the proposed BALI fields and SCPA model. From the experimental results in Tables \ref{tab300w}, \ref{tabaflw}, \ref{tabcofw} and \ref{tabwflw}, we can see that SCPAN (SHN+BALI+SCPA) surpasses the SHN+BALI and the SHN+SCPA, respectively. We can conclude that by integrating the BALI fields and SCPA model into the proposed SCPAN, more effective landmark heatmaps and BALI fields can be produced, which can futher enhance the detection robustness against faces with large poses and heavy occlusions. 

(4) Finally, SCPAN can be boosted with unlabeled face images, therefore, we conduct the following experiments (denoted as semi-SCPAN) by mixing CelebA dataset with 300W, AFLW, COFW and WFLW datasets, respectively. As shown in Tables \ref{tab300w}, \ref{tabaflw}, \ref{tabcofw} and \ref{tabwflw}, semi-SCPAN beats SCPAN on 300W, AFLW, COFW and WFLW datasets, which demonstrates that SCPAN can use unlabeled face images to improve its detection accuracy.
\subsection{Experimental Results and Discussions}
From the experimental results listed in Tables \ref{tab300w}-\ref{tab300VW} and the figures presented in previous subsections, we have the following observations and corresponding analyses.

(1) SCPAN, MHHN \cite{Wan2021RobustFA}, AWing \cite{Wang2019AdaptiveWL} and LUVLi \cite{Kumar2020LUVLiFA} are all heatmap regression-based facial landmark detection methods, while the SCPAN outperforms MHHN, Awing and LUVLi as shown in Tables \ref{tab300w}-\ref{tab300VW}, which indicates that 1) the proposed BALI fields can effectively enhance the facial shape constraints, 2) the proposed SCPA model is able to learn more representative and discriminative features for producing more effective landmark heatmaps, 3) by integrating the proposed BALI fields and SCPA model into a novel SCPAN framework, the detection accuracy can be further improved.

(2) Facial shape constraints are very important for landmark detection task for faces with large poses and heavy occlusions. SCPAN, MHHN \cite{Wan2021RobustFA}, ODN \cite{Zhu2019RobustFL} and OpenPose \cite{Sun2019HighResolutionRF} all aim to construct accurate facial shape constraints to improve their detection accuracy. However, from the experimental results in Tables \ref{tab300w}-\ref{tab300VW}, we can see that SCPAN outperforms the other methods for faces with large poses and complicated occlusions, which verifies that boundary and field constraints introduced by SCPAN can effectively and precisely model the spatial relationships among landmarks.

(3) Compared to other semi-supervised facial landmark detection methods including Honari et al. \cite{Honari2018ImprovingLL}, SBR \cite{Dong2018SupervisionbyRegistrationAU}, TS \cite{Dong2019TeacherSS} and STKI \cite{Zhu2020SpatialTemporalKI}, semi-SCPAN outperforms them, which indicates that more effective facial prior knowledge can be learned by semi-SCPAN for achieving more robust and precise landmark detection. 
\subsection{Weakness of the SCPAN}
\textbf{Occlusion problems. }Our proposed SCPAN can outperform the state-of-the-art methods for faces in normal circumstances and large poses. However, for heavily occluded faces (e.g., hair, cup and microphone as shown in Fig. \ref{ger_field}), the field information becomes less accurate as the area around the landmarks may contain a lot of noise. But this situation can be alleviated by using the boundary heatmaps or larger field region.

\textbf{The inference speed. }The baseline SHN can achieve 100FPS in a single Telsa V100 GPU, while our proposed method needs to use both heatmap and field information to predict landmarks. Moreover, the size of the field region also affects the reference speed. When we use the information of the 7x7 field region, our reference speed will drop to 40 FPS. If we use multiple GPUs, the reference will speed up.
\section{Conclusion}
Robust and precise facial landmark detection is still a very challenging topic due to inaccurate facial shape constraints modeling and insufficient labeled training samples. In this work, we present a SCPAN method to address these problems by seamlessly integrating the BALI fields and SCPA model in a semi-supervised framework. It is shown that the proposed BALI fields can effectively model the spatial relationships among landmarks and the SCPA model can learn more representative and discriminative features for producing more accurate landmark heatmaps and BALI fields, which help achieve more robust and precise facial landmark detection. Moreover, SCPAN can use unlabeled face datasets to further boost its detection accuracy, which effectively reduces the dependence on labeled datasets. Experimental results on challenging benchmark datasets demonstrate that the proposed SCPAN outperforms state-of-the-art methods. It can also be found from the experiment that landmark heatmaps, boundary heatmaps and landmark intensity fields can complement and enhance each other, which further improves the detection accuracy. In the future, we plan to further reduce the dependence of detection accuracy on label information and extend our model to other related topics, such as human pose estimation and hand pose estimation.

%


\ifCLASSOPTIONcaptionsoff
  \newpage
\fi


%


\bibliographystyle{IEEEtran}
\bibliography{IEEEabrv,IEEEexample}
\end{document}